\documentclass[letterpaper, 10 pt, conference]{ieeeconf}  %

\IEEEoverridecommandlockouts                              %

\usepackage{hyperref}
\usepackage{booktabs}       %
\usepackage{amsfonts}       %
\usepackage{nicefrac}       %
\usepackage{microtype}      %
\usepackage{xcolor}         %

\usepackage{graphicx}
\usepackage{csquotes}
\usepackage{multirow}
\usepackage{array}
\usepackage[table]{xcolor}
\usepackage{makecell}
\usepackage{graphbox}
\usepackage{amsmath}
\usepackage{cleveref}
\usepackage{caption}
\usepackage{subcaption}
\usepackage[export]{adjustbox}
\usepackage{cuted}

\usepackage{xspace}

\newcommand{\boldparagraph}[1]{\vspace{0pt}\noindent{\bf #1\;}}

\newcommand{\Ourname}{Spade-SLAM\xspace}

\def\ourtitle{Spade-SLAM: Sparse Correspondence Anchors for\\ Dense Self-Supervised Visual SLAM}
\title{\LARGE \bf \ourtitle}

\author{%
Julian Philipp Schmitt$^{1,2}$,
Weirong Chen$^{1,2,*}$,
Ganlin Zhang$^{1,2}$,
Felix Wimbauer$^{1,2}$,
Daniel Cremers$^{1,2}$%
\thanks{This work has been submitted to the IEEE for possible publication. Copyright may be transferred without notice, after which this version may no longer be accessible.}%
\thanks{This work was supported by the ERC Advanced Grant ``SIMULACRON'' (agreement \#884679), the GNI Project ``AI4Twinning'', and the DFG project CR 250/26-1 ``4DYoutube''.}%
\thanks{$^{1}$All authors are with Technical University of Munich, Munich, Germany.}%
\thanks{$^{2}$All authors are with Munich Center for Machine Learning (MCML), Munich, Germany.}%
\thanks{$^{*}$Corresponding author.}%
}

\begin{document}

\maketitle
\thispagestyle{empty}
\pagestyle{empty}

\begin{strip}
    \centering
    \vspace{-5em}

    \captionsetup{type=figure}
    \includegraphics[width=0.95\textwidth]{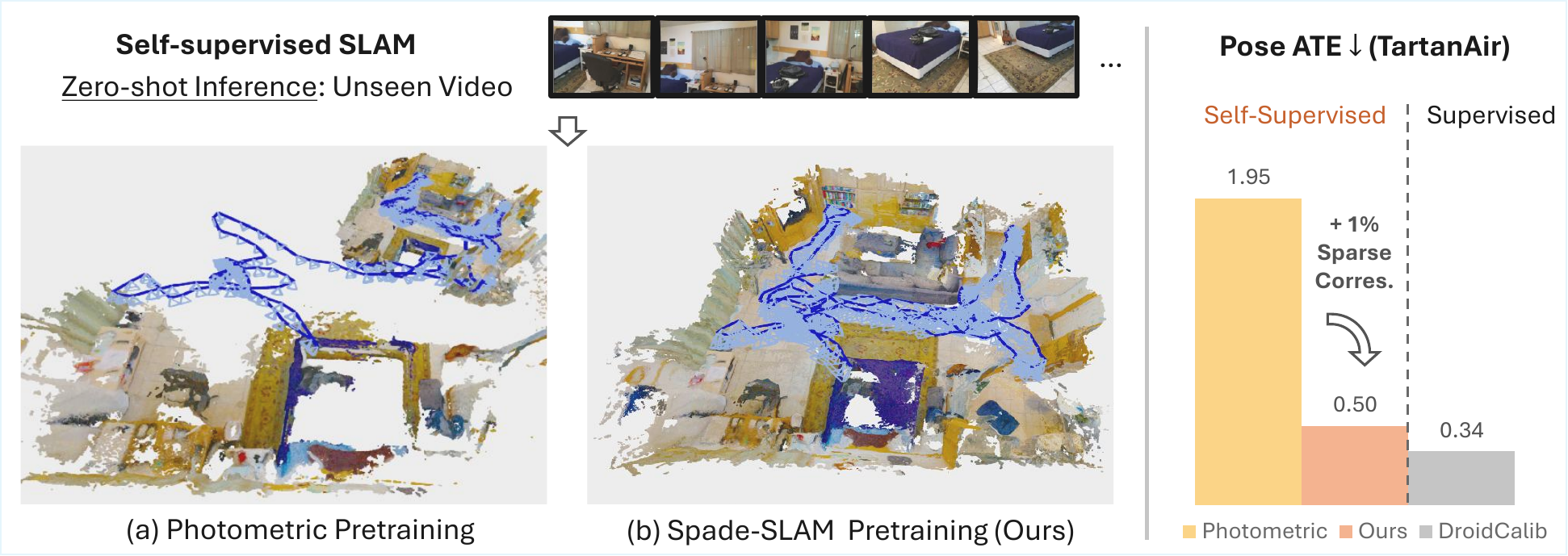}

    \caption{
        \Ourname is a dense visual SLAM system trained via self-supervision
        from only a photometric loss and sparse correspondences.
        Despite requiring no ground-truth labels, it rivals the performance
        of densely supervised approaches such as
        DroidCalib~\cite{hagemann2023droidcalib} on the challenging
        TartanAir~\cite{wang2020tartanair} benchmark in a zero-shot setting.
    }
    \label{fig:teaser}

\end{strip}

\begin{abstract}

We introduce \Ourname, a self-supervised dense visual SLAM framework that
substantially narrows the long-standing performance gap to supervised methods in a
zero-shot setting. By demonstrating that high-fidelity tracking can be achieved
without explicit ground-truth pose and depth labels, our approach offers a
highly effective and label-efficient alternative to fully supervised
pretraining. Our core contribution is a complementary sparse-to-dense training
regime: While a warping-based photometric loss provides a dense training signal,
we stabilize the optimization and suppress drift in ambiguous regions using a
sparse correspondence loss derived from image feature matches. Extensive
experiments demonstrate that \Ourname significantly improves robustness and
accuracy compared to existing self-supervised baselines, closely rivaling SLAM
systems trained with explicit pose and depth supervision. 
The code will be made publicly available at
\url{https://github.com/JulianPSchmitt/Spade-SLAM}.
\end{abstract}

\section{Introduction}
\label{sec:intro}

Recent years have witnessed rapid progress in dense visual simultaneous localization and mapping (SLAM), driven by the growing demand for reliable 3D perception in robotics~\cite{Schmid-RSS24-Khronos}, augmented reality~\cite{usenko2019basalt}, and autonomous navigation~\cite{mur2017orbslam2}. Unlike traditional sparse methods that rely on discrete feature keypoints~\cite{klein2007ptam,mur2017orbslam2}, dense SLAM systems reconstruct detailed geometry, enabling richer spatial understanding for downstream tasks~\cite{newcombe2011dtam}. While classical geometry-based methods laid strong foundations, recent advances increasingly leverage deep learning~\cite{teed2021DROID,hagemann2023droidcalib,murai2025mast3rslam} to improve robustness in texture-less regions, under motion blur, and across challenging lighting conditions, pushing dense SLAM closer to real-world deployment.

Despite their strong performance, state-of-the-art
learning-based SLAM systems rely heavily on supervised training with
ground-truth poses and dense depth annotations~\cite{teed2021DROID}. Acquiring
such comprehensive ground truth is costly and often restricted to synthetic
environments or specialized capture setups, making label-efficient alternatives
highly desirable. Self-supervised learning offers a compelling direction by
typically replacing explicit labels with photometric warping as a dense training
signal~\cite{klodt2018supervising, engel2014lsd, sun2023dynamo}. However,
photometric consistency can be a fragile assumption: it breaks down under
illumination changes, non-Lambertian surfaces, and in texture-less regions.
Consequently, pure photometric methods have struggled to match the accuracy and
robustness of their supervised counterparts, leaving a performance gap that
limits their practical utility~\cite{yu2022d3flowslam}.

In this work, we aim to narrow this gap by demonstrating that self-supervised
learning can achieve performance competitive with fully supervised baselines,
even when constrained to identical training data volumes. We introduce \Ourname,
a self-supervised SLAM framework built upon a complementary sparse-to-dense
training regime. Our key insight is that while dense photometric alignment can
be prone to local minima, sparse point correspondences provide reliable and
illumination-robust geometric cues. By incorporating a sparse correspondences
reprojection loss alongside a robust brightness-aligned photometric objective,
we introduce strong geometric anchors that stabilize optimization and suppress
drift in ambiguous regions, without the need for ground-truth camera poses or
dense depth supervision.

Importantly, our approach enables training with sparse (pseudo-)label supervision
to achieve accurate dense predictions (see~\Cref{fig:teaser}). By coupling dense
photometric supervision with a minimal set of high-confidence point correspondences (less
than $\sim 1\%$ of total image pixels), which can be easily obtained using
classical or learning-based matchers, we significantly advance self-supervised
dense SLAM. This strategy substantially outperforms prior self-supervised
baselines, reaching zero-shot performance comparable to fully supervised
methods. Furthermore, our formulation is self-calibrating: by jointly optimizing
unknown camera focal lengths alongside pose and depth via bundle adjustment,
\Ourname naturally supports adaptation on unseen, uncalibrated videos
to further enhance tracking and reconstruction.

In summary, our contributions include: 
\begin{itemize}%
\item We boost \textbf{dense photometric} self-supervision with off-the-shelf \textbf{sparse geometric} constraints that stabilize optimization and enable accurate dense reconstruction, substantially narrowing the gap to supervised SLAM.
\item We introduce a self-calibrating training framework that jointly optimizes camera focal length, pose, and depth, enabling \textbf{self-supervised} learning without requiring dense ground-truth annotations.
\item Extensive experiments demonstrate that our method significantly outperforms prior self-supervised approaches and achieves performance \textbf{close to fully supervised baselines} in camera tracking and dense 3D reconstruction.
\end{itemize}

\section{Related work}

\subsection{Classical and supervised visual SLAM}

Visual SLAM jointly recovers camera poses and 3D map from RGB video input.
Classical methods fall into two main categories: indirect (feature-based) and
direct (photometric-based) approaches. Indirect systems such as
ORB-SLAM~\cite{mur2017orbslam2, campos2021orbslam3} estimate structure and
motion from sparse hand-crafted keypoint matches, offering efficient and robust
tracking but discarding dense image information and degrading in texture-poor or
repetitive scenes. Direct methods such as DSO~\cite{engel2017dso} and
LSD-SLAM~\cite{engel2014lsd} minimize photometric error over pixel intensities,
but their reliance on brightness constancy makes them sensitive to distortions
such as auto-exposure changes and rolling shutter.

Recent advances have shifted toward learning-based SLAM systems that leverage
data-driven priors~\cite{teed2020deepv2d, yang2020d3vo, teed2021DROID,
hagemann2023droidcalib}. Early works such as DeepV2D~\cite{teed2020deepv2d}
introduced differentiable geometric modules that alternate between depth and
pose refinement, culminating in DROID-SLAM~\cite{teed2021DROID}, which combines
a RAFT-based~\cite{teed2020raft} optical flow network with differentiable Dense
Bundle Adjustment (DBA). More recently, MASt3R-SLAM~\cite{murai2025mast3rslam}
and VGGT-SLAM~\cite{vggtslam2025} extend foundation models that jointly regress
dense geometry, intrinsics, and poses~\cite{liu2025slam3r,zhang2025vistaslam}
into full SLAM pipelines through windowed processing and global optimization.
While large-scale supervised pretraining provides strong priors for
reconstruction and localization, it relies on costly large-scale ground-truth
depth and pose annotations, motivating self-supervised alternatives that can
exploit unlabeled video.

\subsection{Self-supervised learning in SLAM}

To remove dependence on ground-truth labels, self-supervised learning exploits
photometric consistency: under correct geometry and relative pose, one frame
should be reconstructible from another via differentiable warping. Early works
applied this principle to optical
flow~\cite{wang2018occlusion,liu2019selflow,stone2021smurf}, with approaches
such as SMURF~\cite{stone2021smurf} adapting RAFT~\cite{teed2020raft} to
self-supervised training. In rigid scenes, optical flow can be interpreted as
the 2D projection of 3D motion induced by camera pose and scene depth, enabling
the same photometric objective to supervise depth and pose
networks~\cite{zhou2017sfmlearner,zhao2020towards,godard2019monodepth2}. This
paradigm was popularized by SfMLearner~\cite{zhou2017sfmlearner} and refined by
Monodepth2~\cite{godard2019monodepth2} and Dynamo-Depth~\cite{sun2023dynamo},
which strengthen photometric view synthesis with improved consistency modeling
and motion-aware masking while remaining fundamentally pairwise.

Beyond two-view settings, several methods enforce multi-frame consistency and
jointly optimize geometry and camera motion within SLAM-like frameworks.
D$^3$FlowSLAM~\cite{yu2022d3flowslam} combines flow estimation with bundle
adjustment to decouple static and dynamic motion. However, relying mainly on
photometric supervision yields limited robustness in challenging scenarios when
fine-tuning on evaluation-domain data is omitted. Recent transformer-based
feed-forward models such as E-Rayzer~\cite{zhao2025erayzer} impose photometric
consistency over predicted geometry and trajectories with novel view synthesis.
Meanwhile, distillation-based methods such as AnyCam~\cite{wimbauer2025anycam}
do not rely on photometric consistency, but instead obtain supervision from
pretrained dense optical-flow and depth teachers. Bridging these paradigms, our
method enhances raw photometric consistency using lightweight sparse feature
matching, entirely bypassing the need for dense structural priors.

\begin{figure*}[t]
    \centering
        \includegraphics[width=\textwidth, trim={0cm 5cm 0.5cm 3cm}, clip]{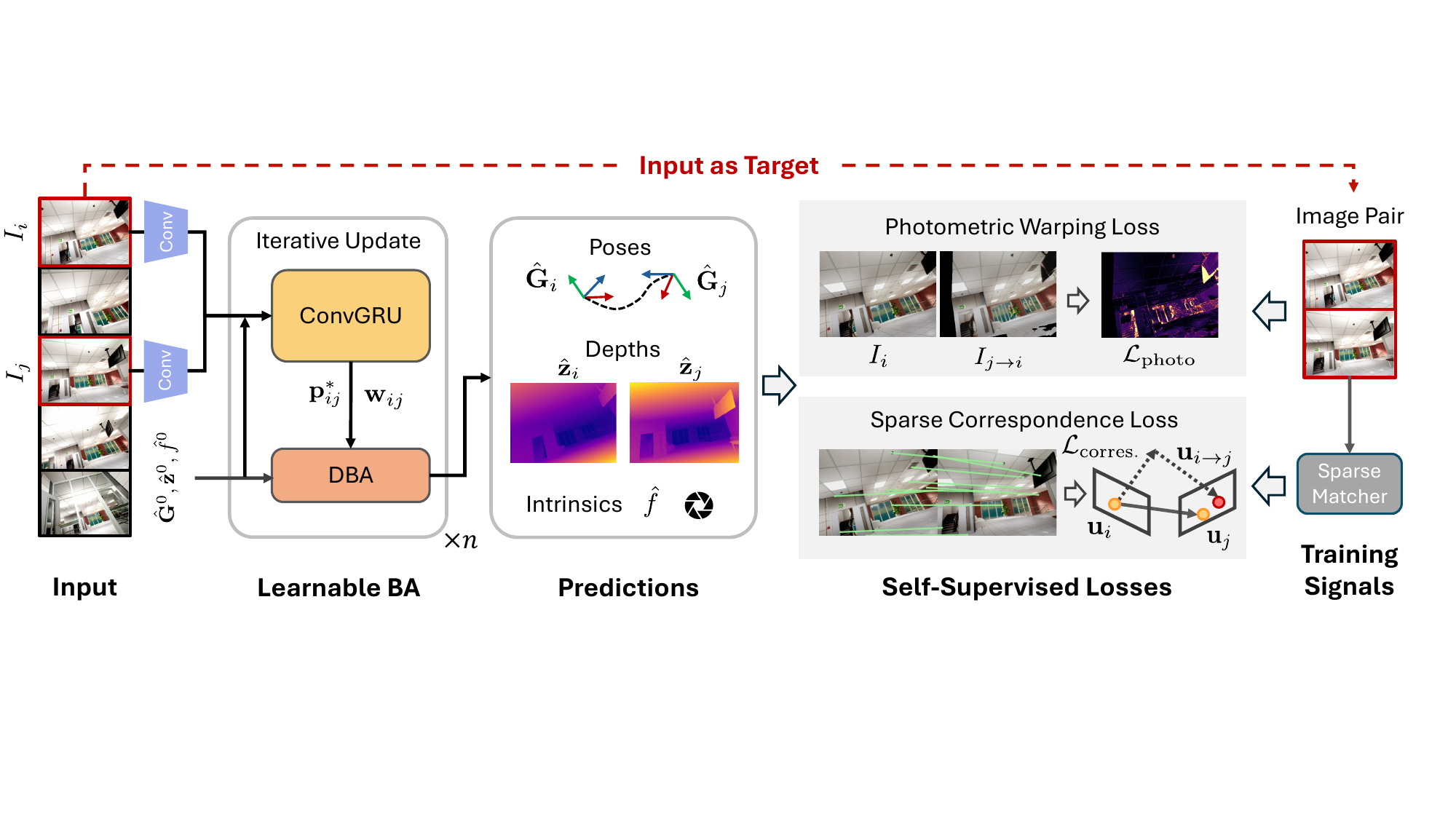}
        \vspace{-0.8em}
    \caption{\textbf{\Ourname pipeline.} 
    Our method jointly optimizes camera poses, dense depth, and a shared focal length over a frame graph $\mathcal{G}=(\mathcal{V}, \mathcal{E})$, where each image pair $(i, j)$ represents an edge for iterative updates. Trained end-to-end via self-supervision, \Ourname leverages photometric warping and sparse correspondence-based reprojection, eliminating the need for ground-truth pose or depth annotations.
    }
    \label{fig:pipeline}
\end{figure*}

\subsection{Sparse anchors for dense reconstruction}

Sparse-to-dense supervision has been explored to mitigate photometric ambiguities in monocular depth estimation, where a small number of reliable anchors (e.g., LiDAR) effectively regularize dense predictions~\cite{kuznietsov2017semisupervised,loquercio2020sparsedepth}.
Similarly, neural scene representation methods often leverage SfM-derived sparse point clouds to initialize or constrain NeRF and 3D Gaussian Splatting~\cite{deng2022depth,roessle2022dense,kerbl202333dgs}. In SLAM, however, this paradigm is less explored: prior methods interpolate dense flow from sparse matches but do not enable end-to-end SLAM training~\cite{revaud2015epicflow, wulff2015efficient}. In contrast, we directly impose lightweight 2D correspondence constraints within a self-supervised SLAM framework, without explicit sparse 3D reconstruction and with preprocessing required only during training.

\section{Methodology} \label{sec:methodology}

Given a casual, uncalibrated RGB video stream $\mathcal{V} = \{I_i \in
\mathbb{R}^{H \times W \times 3}\}_{i=1}^N$, we aim to recover a shared focal length
$f$, the camera poses $\{\mathbf{G}_i \in SE(3)\}_{i=1}^N$, and the 
per-view dense depth maps $\{\mathbf{Z}_i \in \mathbb{R}^{H \times
W}\}_{i=1}^N$ in a self-supervised manner.
However, purely photometric supervision often fails to achieve competitive performance, as the underlying geometry optimization is severely ill-posed under ambiguous visual signals. To address this limitation, we augment photometric supervision with sparse geometric constraints. We demonstrate that injecting a minimal set ($\sim 1\%$ of total image pixels) of high-confidence correspondences is sufficient to stabilize the optimization. When combined with photometric loss, these sparse anchors further enable accurate camera tracking and dense reconstruction without requiring ground-truth pose or depth supervision. Our pipeline is shown in~\Cref{fig:pipeline}.

\subsection{Preliminary: differentiable SLAM framework}\label{sec:deep-visual-slam}

We adopt the differentiable SLAM framework of DROID-SLAM~\cite{teed2021DROID} as our backbone, which couples dense optical flow estimation with iterative bundle adjustment. The system maintains a global geometric state consisting of per-frame camera poses 
$\hat{\mathbf{G}}_i \in SE(3)$ and low-resolution depth maps 
$\hat{\mathbf{z}}_i \in \mathbb{R}^{\frac{H}{8} \times \frac{W}{8}}$, 
defined over a dynamically constructed frame graph 
$\mathcal{G} = (\mathcal{V}, \mathcal{E})$ that encodes co-visibility between frames. Full-resolution dense depth $\hat{\mathbf{Z}}_i \in \mathbb{R}^{H \times W}$ is subsequently recovered by learned convex upsampling.

At each iteration, a recurrent update operator refines this state in two steps. 
First, a RAFT-style ConvGRU~\cite{teed2020raft} module predicts dense correspondence corrections 
$\boldsymbol{\delta}_{ij}$ and confidence weights $\mathbf{w}_{ij}$ 
for each connected image pair $(i,j) \in \mathcal{E}$. 
These corrections are applied to the reprojected pixels induced by the current pose–depth estimates. Specifically, for pixel coordinates $\mathbf{p}_i$ in frame $I_i$, 
the induced projection into frame $I_j$ is given by
\begin{align}
    \mathbf{p}_{ij} = 
    \Pi_{f} \left( \hat{\mathbf{G}}_{ij} 
    \circ \Pi_{f}^{-1}(\mathbf{p}_i, \hat{\mathbf{z}}_i) \right),
    \label{eq:induced_flow}
\end{align}
where $\Pi_f$ and $\Pi_f^{-1}$ denote projection and back-projection under focal 
length $f$, and $\hat{\mathbf{G}}_{ij} = \hat{\mathbf{G}}_j \hat{\mathbf{G}}_i^{-1}$ 
is the relative pose. The predicted correspondences are 
$\mathbf{p}_{ij}^* = \mathbf{p}_{ij} + \boldsymbol{\delta}_{ij}$.

Second, these predicted correspondences are passed to the differentiable Dense Bundle 
Adjustment (DBA) layer, which performs a weighted Gauss–Newton update of the geometric state $\mathbf{\Theta} = (\hat{\mathbf{G}}, \hat{\mathbf{z}})$ by minimizing the reprojection residual:
\begin{align}
\min_{\mathbf{\Theta}} 
\sum_{(i,j) \in \mathcal{E}} 
\left\| \mathbf{p}_{ij}^* - \mathbf{p}_{ij} \right\|_{\Sigma_{ij}}^2, 
\qquad 
\Sigma_{ij} = \mathrm{diag}\, \mathbf{w}_{ij}.
\label{eq:dba}
\end{align} 
This optimization loop forms the geometric core of our system. In the following sections, we reformulate its supervision with the self-supervised objectives that enable learning without explicit pose and depth annotations.

\subsection{Brightness-aligned photometric supervision} \label{sec:photometric-supervision}

We adopt the widely-used photometric consistency loss as the primary supervision
signal, which encourages the predicted geometry to warp one image into
another~\cite{engel2017dso,engel2014lsd,zhou2017sfmlearner}. Under the correct
pose–depth estimates, one frame should be reconstructible from another via
differentiable warping with the induced reprojection. The deviations from
brightness constancy therefore provide a supervision signal for both camera
poses and scene structures.

Given the current pose–depth estimates $(\hat{\mathbf{G}}_{ij},
\hat{\mathbf{Z}}_i)$, pixel coordinates from frame $I_i$ are projected into
frame $I_j$ using the projection model in \eqref{eq:induced_flow}. Bilinear
sampling of pixel intensities then yields a synthesized view
\begin{align}
I_{j \to i} =
I_j \left(
\Pi_{{f}}\big(
\hat{\mathbf{G}}_{ij}
\circ
\Pi_{{f}}^{-1}(\mathbf{p}_i, \hat{\mathbf{Z}}_i)
\big)
\right).
\end{align}

In practice, directly minimizing the photometric discrepancy between $I_i$ and
$I_{j \to i}$ is brittle as the brightness constancy assumption is frequently
violated due to auto-exposure adjustments, sensor response variations, and
dynamic illumination. Such radiometric inconsistencies introduce spurious
photometric residuals that can dominate the optimization and misguide the
geometric updates. Inspired by the idea of photometric
calibration~\cite{engel2017dso}, we model inter-frame intensity changes using a
global affine transformation to improve robustness. Specifically, we estimate
scalar parameters $a_{ij}, b_{ij} \in \mathbb{R}$ that minimize
\begin{align}
    \min_{a_{ij}, b_{ij}} 
    \left\lVert 
    I_i - (a_{ij} I_{j \to i} + b_{ij}) 
    \right\rVert_2^2.
\end{align}
This least-squares problem admits the closed-form solution 
\begin{align}
    a_{ij} = \frac{\text{cov}(I_i, I_{j \to i})}{\text{var}(I_{j \to i})}, \qquad
    b_{ij} = \mu_{I_i} - a_{ij}\mu_{I_{j \to i}},
\end{align}
where $\mu$, $\text{var}(\cdot)$, and
$\text{cov}(\cdot,\cdot)$ denote pixel-wise statistics. The aligned image is
given by $\tilde{I}_{j \to i} = a_{ij} I_{j \to i} + b_{ij}$ and importantly
$a_{ij}, b_{ij}$ are treated as constants during backpropagation.

The final photometric objective combines SSIM~\cite{wang2004ssim} and pixel-wise
residuals:
\begin{align}
    \mathcal{L}_{\text{photo}}
    = \frac{1}{|\mathcal{E}|}
    \sum_{(i,j)\in\mathcal{E}}
    \Big(
    &\frac{\alpha}{2}
    \big(1-\operatorname{SSIM}(I_i,\tilde I_{j\to i})\big)
    \notag\\[-0.5em]
    &+(1-\alpha)
    \lVert I_i-\tilde I_{j\to i}\rVert_1
    \Big).
    \label{eq:photo_loss}
\end{align}
with $\alpha = 0.85$. By compensating for global brightness variations, this alignment stabilizes photometric supervision and enables effective self-supervised training on challenging video sequences.

\subsection{Sparse correspondence anchoring} 
\label{sec:sparse-flow-supervision}

While dense photometric supervision provides broad spatial coverage, it relies
on brightness constancy and is therefore sensitive to illumination changes and
non-Lambertian effects, and can struggle under large baseline motion. In such
scenarios, different pose–depth configurations can produce similar photometric
residuals, making the underlying optimization ill-posed and prone to local
minima.

We address this by injecting a minimal set of high-confidence sparse
correspondences as explicit geometric anchors. Importantly, these anchors
operate directly on the projection model used in DBA and therefore constrain the
same pose–depth state variables optimized in \eqref{eq:dba}. Despite covering
less than 1\% of image pixels in practice, we find that such sparse constraints
are sufficient to regularize the optimization and significantly improve
robustness.

A key benefit of our approach is the ease of obtaining sparse correspondences: they are readily accessible using classical feature matching with keypoint extractors or modern learning-based methods~\cite{lowe2004sift,detone2018superpoint,lindenberger2023lightglue}. For each image pair $(i,j)\in\mathcal E$, we denote the $K_{ij}$ validated correspondences as
\begin{equation}
\mathcal M_{ij}
=
\left\{
(\mathbf u_i^k,\mathbf u_j^k,w_{ij}^k)
\right\}_{k=1}^{K_{ij}},
\end{equation}
where $\mathbf u_i^k$ and $\mathbf u_j^k$ are matched image coordinates and
$w_{ij}^k$ denotes a confidence weight from the feature matching process. For
each source keypoint $\mathbf u_i^k$, we denote by $\mathbf u_{i\to j}^k$ its
induced projection into frame $I_j$ under the current geometric state, computed
analogously to the reprojection in \eqref{eq:induced_flow}. We then minimize
the confidence-weighted reprojection error
\begin{align}
\mathcal L_{\text{corres.}}
&=
\frac{1}{|\mathcal E|}
\sum_{(i,j)\in\mathcal E}
\frac{1}{W_{ij}}
\sum_{k=1}^{K_{ij}}
w_{ij}^k \
\lVert
\mathbf u_{i\to j}^k-\mathbf u_j^k
\rVert_\delta ,
\label{eq:sparse_flow_loss}
\end{align}
where $\lVert\cdot\rVert_\delta$ denotes the Huber norm with $\delta=1$, and
$
W_{ij}
=
\sum_{k=1}^{K_{ij}} w_{ij}^k
$
normalizes the contribution of each image pair, preventing pairs with more
correspondences from dominating the gradients. By directly supervising the
projection geometry at sparse but reliable locations, this auxiliary objective
provides structural anchors that complement photometric alignment.

\subsection{Self-calibrated training with unknown intrinsics} \label{sec:self-calib-training}

We further extend our framework to a self-calibrating setting, eliminating the
need for ground-truth camera intrinsics during training. In contrast to prior
approaches~\cite{hagemann2023droidcalib, teed2021DROID} that rely on
pre-calibrated intrinsics during training, we incorporate the focal length as an
additional optimization variable in the geometric state, i.e. $\mathbf{\Theta} =
(\hat{\mathbf{G}}, \hat{\mathbf{z}}, \hat{f})$. The focal length is initialized
with a coarse resolution-based prior, $\hat{f}_{\text{init}} = \frac{H + W}{2}$,
and subsequently optimized via photometric and sparse geometric supervision.
This relaxes the need for pre-calibration and enables test-time adaptation on
uncalibrated videos.

\boldparagraph{Optimization objective}
We train the model end-to-end on monocular video subsequences.
Following~\cite{hagemann2023droidcalib, teed2021DROID, yu2022d3flowslam}, we
sample 6-frame clips and unroll the recurrent update operator for $n$
iterations, applying deep supervision to intermediate predictions with
exponentially increasing weights ($\gamma=0.9$). The training objective combines
the dense photometric alignment loss, $\mathcal{L}_{\text{photo}}$, with the sparse
geometric reprojection loss, $\mathcal{L}_{\text{corres.}}$. Together, these terms
define a geometry-anchored self-supervised bundle adjustment objective in which
dense photometric coverage and sparse correspondence constraints jointly shape
the optimization landscape.

\section{Experiments} \label{sec:experiments}

\begin{table*}[t]
\centering
\caption{\textbf{TartanAir~\cite{wang2020tartanair} benchmark (hard split): ATE RMSE
[m].} The $\dagger$ symbol indicates methods using fine-tuning on the
evaluation data. - means lack of results. 
\Ourname performs the best on average across the self-supervised methods 
and performs close to supervised methods.
}
\label{tab:tartanair_results}

\renewcommand{\arraystretch}{0.9} 

\resizebox{0.9\textwidth}{!}{%
\setlength{\aboverulesep}{1pt}
\setlength{\belowrulesep}{1pt}
\scriptsize
\setlength{\tabcolsep}{3pt}

\begin{tabular}{llccccccccc} 
\toprule

\multicolumn{2}{c}{\textbf{Method}} 
& \texttt{MH000} 
& \texttt{MH001} 
& \texttt{MH002} 
& \texttt{MH003} 
& \texttt{MH004} 
& \texttt{MH005} 
& \texttt{MH006} 
& \texttt{MH007} 
& \textbf{Avg.} \\

\midrule

\multirow{2}{*}{\rotatebox[origin=c]{90}{\color{gray}\textit{Trad.}}} 
& \color{gray}ORB-SLAM3~\cite{campos2021orbslam3} 
& \color{gray}15.44 
& \color{gray}2.92 
& \color{gray}13.51 
& \color{gray}8.18 
& \color{gray}\textbf{2.59} 
& \color{gray}21.91 
& \color{gray}11.70 
& \color{gray}25.88 
& \color{gray}\textbf{12.77} \\

& \color{gray}DSO~\cite{engel2017dso} 
& \color{gray}\textbf{9.92} 
& \color{gray}\textbf{0.35} 
& \color{gray}\textbf{7.96} 
& \color{gray}\textbf{3.46} 
& \color{gray}- 
& \color{gray}\textbf{12.58} 
& \color{gray}\textbf{8.42} 
& \color{gray}\textbf{7.50} 
& \color{gray}- \\

\cmidrule{1-11} 

\multirow{4}{*}{\rotatebox[origin=c]{90}{\color{gray}\textit{Sup.}}} 
& \color{gray}DeepV2D~\cite{teed2020deepv2d} 
& \color{gray}6.15 
& \color{gray}2.12 
& \color{gray}4.54 
& \color{gray}3.89 
& \color{gray}2.71 
& \color{gray}11.55 
& \color{gray}5.53 
& \color{gray}3.76 
& \color{gray}5.03 \\

& \color{gray}TartanVO~\cite{corl2020tartanvo} 
& \color{gray}4.88 
& \color{gray}0.26 
& \color{gray}2.00 
& \color{gray}0.94 
& \color{gray}\textbf{1.07} 
& \color{gray}3.19 
& \color{gray}1.00 
& \color{gray}2.04 
& \color{gray}1.92 \\

& \color{gray}DROID-SLAM~\cite{teed2021DROID} 
& \color{gray}0.07 
& \color{gray}0.05 
& \color{gray}0.03 
& \color{gray}\textbf{0.02} 
& \color{gray}3.03 
& \color{gray}0.63 
& \color{gray}0.38 
& \color{gray}\textbf{0.07} 
& \color{gray}0.54 \\

& \color{gray}DroidCalib~\cite{hagemann2023droidcalib} 
& \color{gray}\textbf{0.06} 
& \color{gray}\textbf{0.03} 
& \color{gray}\textbf{0.02} 
& \color{gray}\textbf{0.02} 
& \color{gray}1.89 
& \color{gray}\textbf{0.30} 
& \color{gray}\textbf{0.25} 
& \color{gray}0.13 
& \color{gray}\textbf{0.34} \\

\cmidrule{1-11}

\multirow{6}{*}{\rotatebox[origin=c]{90}{\textit{S.-Sup.}}} 
& AnyCam~\cite{wimbauer2025anycam} 
& 14.44 
& 2.89 
& 12.35 
& 2.97 
& 2.77 
& 12.94 
& 7.15 
& 7.99 
& 7.94 \\

& Dynamo-Depth\smash{$^\dagger$}~\cite{sun2023dynamo} 
& 33.36 
& 7.67 
& 15.89 
& 10.13 
& 9.39 
& 19.46 
& 16.94 
& 15.08 
& 15.99 \\

& D\smash{$^3$}FlowSLAM\smash{$^\dagger$}~\cite{yu2022d3flowslam} 
& \cellcolor{orange!25}0.83 
& \cellcolor{yellow!25}0.11 
& \cellcolor{yellow!25}0.19 
& 0.36 
& \cellcolor{yellow!25}2.41 
& 1.62 
& \cellcolor{red!25}\textbf{0.19} 
& 0.58 
& \cellcolor{yellow!25}0.79 \\

& Ours (photo. only) 
& \cellcolor{yellow!25}0.99 
& \cellcolor{orange!25}0.07 
& \cellcolor{orange!25}0.03 
& \cellcolor{orange!25}0.03 
& 3.88 
& \cellcolor{yellow!25}0.99 
& 9.22 
& \cellcolor{yellow!25}0.43 
& 1.95 \\

& Ours (SIFT + NN) 
& 1.21 
& 0.77 
& \cellcolor{red!25}\textbf{0.02} 
& \cellcolor{red!25}\textbf{0.01} 
& \cellcolor{red!25}\textbf{1.92} 
& \cellcolor{red!25}\textbf{0.82} 
& \cellcolor{orange!25}0.35 
& \cellcolor{red!25}\textbf{0.16} 
& \cellcolor{orange!25}0.66 \\

& \textbf{Ours} 
& \cellcolor{red!25}\textbf{0.36} 
& \cellcolor{red!25}\textbf{0.04} 
& \cellcolor{orange!25}0.03 
& \cellcolor{yellow!25}0.05 
& \cellcolor{orange!25}2.04 
& \cellcolor{orange!25}0.88 
& \cellcolor{yellow!25}0.38 
& \cellcolor{orange!25}0.24 
& \cellcolor{red!25}\textbf{0.50}\\

\bottomrule
\end{tabular}%
}

\end{table*}

\begin{table*}[t]
\centering
\caption{\textbf{EuRoC MAV~\cite{burri2016euroc} benchmark: ATE RMSE [m].} The $\dagger$
symbol indicates methods using fine-tuning on the evaluation data. X
means the system has failed here. - means lack of results. \Ourname outperforms
other self-supervised methods on average and matches
DROID-SLAM~\cite{teed2021DROID}.}
\label{tab:euroc_results}
\renewcommand{\arraystretch}{0.9}

\resizebox{0.9\textwidth}{!}{%
\footnotesize
\setlength{\aboverulesep}{1pt}
\setlength{\belowrulesep}{1pt}
\setlength{\tabcolsep}{3pt}
\begin{tabular}{llcccccccccccc}
\toprule
\multicolumn{2}{c}{\textbf{Method}} & \texttt{MH01} & \texttt{MH02} & \texttt{MH03} & \texttt{MH04} & \texttt{MH05} & \texttt{V101} & \texttt{V102} & \texttt{V103} & \texttt{V201} & \texttt{V202} & \texttt{V203} & \textbf{Avg.} \\
\midrule

\multirow{2}{*}{\rotatebox[origin=c]{90}{\color{gray}\textit{Trad.}}} 
 & \color{gray}ORB-SLAM3~\cite{campos2021orbslam3} & \color{gray}\textbf{0.016} & \color{gray}\textbf{0.027} & \color{gray}\textbf{0.028} & \color{gray}\textbf{0.138} & \color{gray}\textbf{0.072} & \color{gray}\textbf{0.033} & \color{gray}\textbf{0.015} & \color{gray}\textbf{0.033} & \color{gray}\textbf{0.023} & \color{gray}\textbf{0.029} & \color{gray}X & \color{gray}- \\
 & \color{gray}DSO~\cite{engel2017dso} & \color{gray}0.046 & \color{gray}0.046 & \color{gray}0.172 & \color{gray}3.810 & \color{gray}0.110 & \color{gray}0.089 & \color{gray}0.107 & \color{gray}0.903 & \color{gray}0.044 & \color{gray}0.132 & \color{gray}\textbf{1.152} & \color{gray}\textbf{0.601} \\
\midrule

\multirow{5}{*}{\rotatebox[origin=c]{90}{\color{gray}\textit{Sup.}}} 
 & \color{gray}DeepV2D~\cite{teed2020deepv2d} & \color{gray}0.739 & \color{gray}1.144 & \color{gray}0.752 & \color{gray}1.492 & \color{gray}1.567 & \color{gray}0.981 & \color{gray}0.801 & \color{gray}1.570 & \color{gray}0.290 & \color{gray}2.202 & \color{gray}2.743 & \color{gray}1.298 \\
 & \color{gray}TartanVO~\cite{corl2020tartanvo} & \color{gray}0.639 & \color{gray}0.325 & \color{gray}0.550 & \color{gray}1.153 & \color{gray}1.021 & \color{gray}0.447 & \color{gray}0.389 & \color{gray}0.622 & \color{gray}0.433 & \color{gray}0.749 & \color{gray}1.152 & \color{gray}0.680 \\
 & \color{gray}MASt3R-SLAM~\cite{murai2025mast3rslam} & \color{gray}0.023 & \color{gray}0.017 & \color{gray}0.057 & \color{gray}0.113 & \color{gray}0.067 & \color{gray}0.040 & \color{gray}0.019 & \color{gray}0.027 & \color{gray}0.020 & \color{gray}0.025 & \color{gray}0.043 & \color{gray}0.041 \\
 & \color{gray}DROID-SLAM~\cite{teed2021DROID} & \color{gray}\textbf{0.011} & \color{gray}\textbf{0.011} & \color{gray}\textbf{0.022} & \color{gray}0.043 & \color{gray}\textbf{0.041} & \color{gray}\textbf{0.036} & \color{gray}0.012 & \color{gray}\textbf{0.018} & \color{gray}\textbf{0.015} & \color{gray}0.010 & \color{gray}\textbf{0.013} & \color{gray}\textbf{0.021} \\
 & \color{gray}DroidCalib~\cite{hagemann2023droidcalib} & \color{gray}\textbf{0.011} & \color{gray}0.012 & \color{gray}\textbf{0.022} & \color{gray}\textbf{0.040} & \color{gray}\textbf{0.041} & \color{gray}\textbf{0.036} & \color{gray}\textbf{0.011} & \color{gray}0.019 & \color{gray}\textbf{0.015} & \color{gray}\textbf{0.009} & \color{gray}0.014 & \color{gray}\textbf{0.021} \\
\midrule

\multirow{6}{*}{\rotatebox[origin=c]{90}{\textit{S.-Sup.}}} 
 & AnyCam~\cite{wimbauer2025anycam} 
 & 2.294 
 & 1.967 
 & 3.264 
 & 4.789 
 & 5.433 
 & 1.749 
 & 1.762 
 & 1.427 
 & 2.165 
 & 2.107 
 & 1.923 
 & 2.625 \\

 & Dynamo-Depth$^\dagger$~\cite{sun2023dynamo} 
 & 4.257 
 & 3.773 
 & 3.430 
 & 6.298 
 & 6.460 
 & 1.762 
 & 1.746 
 & 1.480 
 & 2.017 
 & 2.054 
 & 1.906 
 & 3.198 \\

 & D$^3$FlowSLAM$^\dagger$~\cite{yu2022d3flowslam} 
 & \cellcolor{yellow!25}0.164 
 & 0.184 
 & \cellcolor{yellow!25}0.210 
 & 0.243 
 & 0.775 
 & \cellcolor{yellow!25}0.058 
 & \cellcolor{yellow!25}0.161 
 & 1.360 
 & \cellcolor{yellow!25}0.041 
 & \cellcolor{yellow!25}1.751 
 & \cellcolor{yellow!25}1.228 
 & 0.561 \\

 & Ours (photo. only) 
 & \cellcolor{orange!25}0.011 
 & \cellcolor{yellow!25}0.016 
 & \cellcolor{orange!25}0.022 
 & \cellcolor{yellow!25}0.050 
 & \cellcolor{yellow!25}0.050 
 & \cellcolor{red!25}\textbf{0.031} 
 & \cellcolor{red!25}\textbf{0.007} 
 & \cellcolor{yellow!25}0.090 
 & \cellcolor{orange!25}0.014 
 & \cellcolor{orange!25}0.025 
 & \cellcolor{orange!25}0.091 
 & \cellcolor{yellow!25}0.037 \\

 & Ours (SIFT + NN) 
 & \cellcolor{red!25}\textbf{0.009} 
 & \cellcolor{red!25}\textbf{0.012} 
 & \cellcolor{red!25}\textbf{0.021} 
 & \cellcolor{red!25}\textbf{0.044} 
 & \cellcolor{orange!25}0.044 
 & \cellcolor{orange!25}0.032 
 & \cellcolor{red!25}\textbf{0.007} 
 & \cellcolor{orange!25}0.050 
 & \cellcolor{red!25}\textbf{0.013} 
 & \cellcolor{red!25}\textbf{0.009} 
 & \cellcolor{red!25}\textbf{0.014} 
 & \cellcolor{orange!25}0.023 \\

 & \bf Ours 
 & \cellcolor{orange!25}0.011 
 & \cellcolor{orange!25}0.013 
 & \cellcolor{red!25}\textbf{0.021} 
 & \cellcolor{orange!25}0.046 
 & \cellcolor{red!25}\textbf{0.040} 
 & \cellcolor{orange!25}0.032 
 & \cellcolor{orange!25}0.008 
 & \cellcolor{red!25}\textbf{0.026} 
 & \cellcolor{orange!25}0.014 
 & \cellcolor{red!25}\textbf{0.009} 
 & \cellcolor{red!25}\textbf{0.014} 
 & \cellcolor{red!25}\textbf{0.021} \\

\bottomrule
\end{tabular}%
}
\end{table*}

\boldparagraph{Evaluation datasets and metrics}
We evaluate our method on both camera tracking and dense reconstruction. For
camera trajectory estimation, we use the TartanAir~\cite{wang2020tartanair} test
split and additionally report cross-dataset generalization on
EuRoC~\cite{burri2016euroc} and TUM-RGBD~\cite{sturm2012ate}. Camera pose
accuracy is measured by the RMSE of the $Sim(3)$-aligned Absolute Trajectory
Error (ATE)~\cite{sturm2012ate}. For reconstruction quality, we evaluate on
7-Scenes~\cite{shotton20137scenes}, reporting accuracy, completion, and Chamfer
Distance after $Sim(3)$ alignment to the ground-truth
geometry~\cite{liu2025slam3r}. Best results are highlighted as
\colorbox{red!20}{\textbf{first}}, \colorbox{orange!20}{second}, and
\colorbox{yellow!20}{third}.

\boldparagraph{Implementation details}
Following \cite{teed2021DROID,hagemann2023droidcalib}, we train all learnable
components of \Ourname from scratch on monocular sequences from the synthetic
TartanAir~\cite{wang2020tartanair} dataset. Optimization is performed with
Adam~\cite{kingma2015adam} for 250k steps using a batch size of 4 at $384 \times
512$ resolution, an initial learning rate of $2.5 \times 10^{-4}$, and weight
decay of $10^{-5}$. The update operator is unrolled for $n=15$ iterations. We obtain sparse correspondences by extracting and matching up
to 2048 keypoints per image pair, followed by epipolar filtering, resulting in
less than 1\% correspondence coverage. We evaluate two correspondence pipelines:
(1) learned SuperPoint~\cite{detone2018superpoint} keypoints with
LightGlue~\cite{lindenberger2023lightglue} matching, which constitutes our main
model and is denoted \textit{Ours} hereafter; and (2) classical, fully
self-supervised SIFT~\cite{lowe2004sift} keypoints with nearest-neighbor
matching, denoted \textit{Ours (SIFT + NN)}. To quantify the contribution of
sparse correspondence guidance, we further  evaluate a self-supervised variant
trained with pure photometric loss, denoted as \textit{Ours (photo. only)}.
Training takes one week on four NVIDIA L40S GPUs.

\begin{table*}[t]
\centering
\caption{\textbf{TUM-RGBD~\cite{sturm2012ate} benchmark: ATE RMSE [m].} X means the system has failed here. - means lack of results. \Ourname achieves the best ATE on average among self-supervised methods and matches supervised DroidCalib~\cite{hagemann2023droidcalib}.}
\vspace{0.5em}
\label{tab:tum_results}

\renewcommand{\arraystretch}{0.9}

\resizebox{0.9\textwidth}{!}{%
\setlength{\aboverulesep}{1pt}
\setlength{\belowrulesep}{1pt}
\footnotesize
\setlength{\tabcolsep}{5pt}

\begin{tabular}{llcccccccccc}
\toprule

\multicolumn{2}{c}{\textbf{Method}} 
& \texttt{360} 
& \texttt{desk} 
& \texttt{desk2} 
& \texttt{floor} 
& \texttt{plant} 
& \texttt{room} 
& \texttt{rpy} 
& \texttt{teddy} 
& \texttt{xyz} 
& \textbf{Avg.} \\

\midrule

\multirow{2}{*}{\rotatebox[origin=c]{90}{\color{gray}\textit{Trad.}}} 
& \color{gray}ORB-SLAM2~\cite{mur2017orbslam2} 
& \color{gray}X 
& \color{gray}0.071 
& \color{gray}X 
& \color{gray}\textbf{0.023} 
& \color{gray}X 
& \color{gray}X 
& \color{gray}X 
& \color{gray}X 
& \color{gray}0.010 
& \color{gray}- \\

& \color{gray}ORB-SLAM3~\cite{campos2021orbslam3} 
& \color{gray}X 
& \color{gray}\textbf{0.017} 
& \color{gray}\textbf{0.210} 
& \color{gray}X 
& \color{gray}\textbf{0.034} 
& \color{gray}X 
& \color{gray}X 
& \color{gray}X 
& \color{gray}\textbf{0.009} 
& \color{gray}- \\

\midrule 

\multirow{5}{*}{\rotatebox[origin=c]{90}{\color{gray}\textit{Sup.}}} 
& \color{gray}DeepV2D~\cite{teed2020deepv2d} 
& \color{gray}0.243 
& \color{gray}0.166 
& \color{gray}0.379 
& \color{gray}1.653 
& \color{gray}0.203 
& \color{gray}0.246 
& \color{gray}0.105 
& \color{gray}0.316 
& \color{gray}0.064 
& \color{gray}0.375 \\

& \color{gray}TartanVO~\cite{corl2020tartanvo} 
& \color{gray}0.178 
& \color{gray}0.125 
& \color{gray}0.122 
& \color{gray}0.349 
& \color{gray}0.297 
& \color{gray}0.333 
& \color{gray}0.049 
& \color{gray}0.339 
& \color{gray}0.062 
& \color{gray}0.206 \\

& \color{gray}MASt3R-SLAM~\cite{murai2025mast3rslam} 
& \color{gray}\textbf{0.049} 
& \color{gray}\textbf{0.016} 
& \color{gray}\textbf{0.024} 
& \color{gray}0.025 
& \color{gray}0.020 
& \color{gray}0.061 
& \color{gray}0.027 
& \color{gray}0.041 
& \color{gray}\textbf{0.009} 
& \color{gray}\textbf{0.030} \\

& \color{gray}DROID-SLAM~\cite{teed2021DROID} 
& \color{gray}0.069 
& \color{gray}0.017 
& \color{gray}0.025 
& \color{gray}\textbf{0.022} 
& \color{gray}\textbf{0.016} 
& \color{gray}\textbf{0.048} 
& \color{gray}\textbf{0.022} 
& \color{gray}0.037 
& \color{gray}0.010 
& \color{gray}\textbf{0.030} \\

& \color{gray}DroidCalib~\cite{hagemann2023droidcalib} 
& \color{gray}0.071 
& \color{gray}0.017 
& \color{gray}0.028 
& \color{gray}\textbf{0.022} 
& \color{gray}0.095 
& \color{gray}0.069 
& \color{gray}0.023 
& \color{gray}\textbf{0.024} 
& \color{gray}0.034 
& \color{gray}0.043 \\

\midrule 

\multirow{6}{*}{\rotatebox[origin=c]{90}{\textit{S.-Sup.}}}

& AnyCam~\cite{wimbauer2025anycam} 
& \cellcolor{yellow!25}0.165 
& 0.111 
& \cellcolor{yellow!25}0.138 
& 0.547 
& 0.641 
& \cellcolor{yellow!25}0.207 
& 0.040 
& 0.684 
& 0.067 
& 0.289 \\

& Dynamo-Depth~\cite{sun2023dynamo} 
& 0.177 
& 0.838 
& 0.930 
& 0.705 
& 0.663 
& 0.987 
& 0.048 
& 0.868 
& 0.182 
& 0.600 \\

& D\smash{$^3$}FlowSLAM~\cite{yu2022d3flowslam} 
& 0.180 
& \cellcolor{yellow!25}0.019 
& \cellcolor{yellow!25}0.138 
& 0.344 
& \cellcolor{red!25}\textbf{0.019} 
& 0.323 
& 0.044 
& \cellcolor{yellow!25}0.062 
& \cellcolor{yellow!25}0.018 
& \cellcolor{yellow!25}0.127 \\

& Ours (photo. only) 
& \cellcolor{orange!25}0.160 
& \cellcolor{orange!25}0.017 
& \cellcolor{red!25}\textbf{0.026} 
& \cellcolor{yellow!25}0.027 
& 0.699 
& 0.365 
& \cellcolor{red!25}\textbf{0.020} 
& 0.740 
& \cellcolor{orange!25}0.010 
& 0.229 \\

& Ours (SIFT + NN) 
& 0.194 
& \cellcolor{red!25}\textbf{0.016} 
& \cellcolor{red!25}\textbf{0.026} 
& \cellcolor{red!25}\textbf{0.021} 
& \cellcolor{yellow!25}0.090 
& \cellcolor{orange!25}0.083 
& \cellcolor{yellow!25}0.038 
& \cellcolor{red!25}\textbf{0.033} 
& \cellcolor{red!25}\textbf{0.009} 
& \cellcolor{orange!25}0.057 \\

& \textbf{Ours} 
& \cellcolor{red!25}\textbf{0.130} 
& \cellcolor{orange!25}0.017 
& \cellcolor{orange!25}0.029 
& \cellcolor{orange!25}0.022 
& \cellcolor{orange!25}0.059 
& \cellcolor{red!25}\textbf{0.048} 
& \cellcolor{orange!25}0.021 
& \cellcolor{orange!25}0.034 
& \cellcolor{orange!25}0.010 
& \cellcolor{red!25}\textbf{0.041} \\

\bottomrule
\end{tabular}%
}
\end{table*}

\begin{figure*}[t]
    \centering
    \setlength{\tabcolsep}{1pt} 
    \renewcommand{\arraystretch}{0.2} 

    \newcommand{\plotheighttop}{2.4cm}
    \newcommand{\plotheightbottom}{2.65cm}

    \scriptsize
    \resizebox{0.85\linewidth}{!}{%
    \begin{tabular}{ccccc}
        & \large\textbf{\hspace{25pt}DroidCalib} & \large\textbf{AnyCam} & \large\textbf{Ours photo. only} & \large\textbf{\hspace{-25pt}Ours} \\
        
        \raisebox{-0.3cm}{\rotatebox{90}{\large\textbf{room}}}\hspace{5pt} &
        \includegraphics[valign=c, height=\plotheighttop, trim={0cm 1cm 1cm 0cm}, clip]{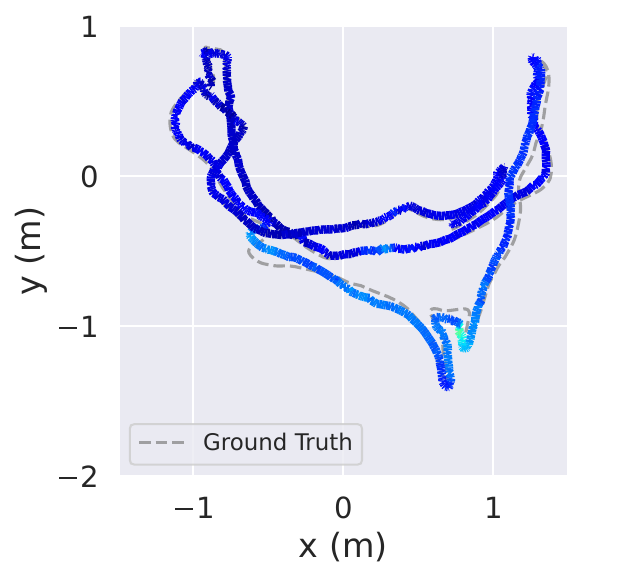} &
        \includegraphics[valign=c, height=\plotheighttop, trim={1.4cm 1cm 1cm 0cm}, clip]{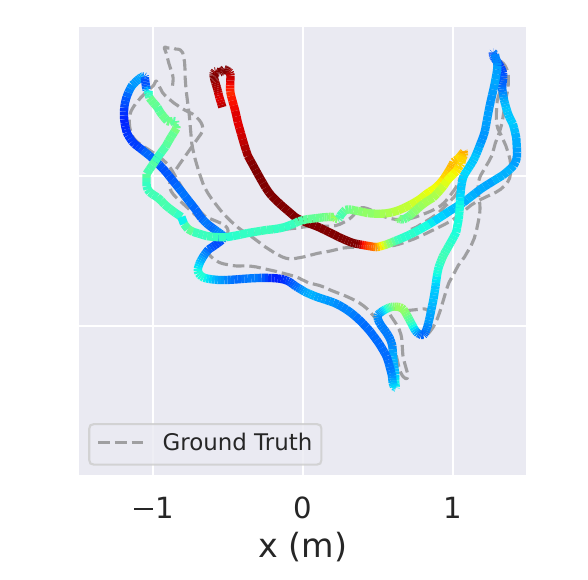} &
        \includegraphics[valign=c, height=\plotheighttop, trim={1.4cm 1cm 1cm 0cm}, clip]{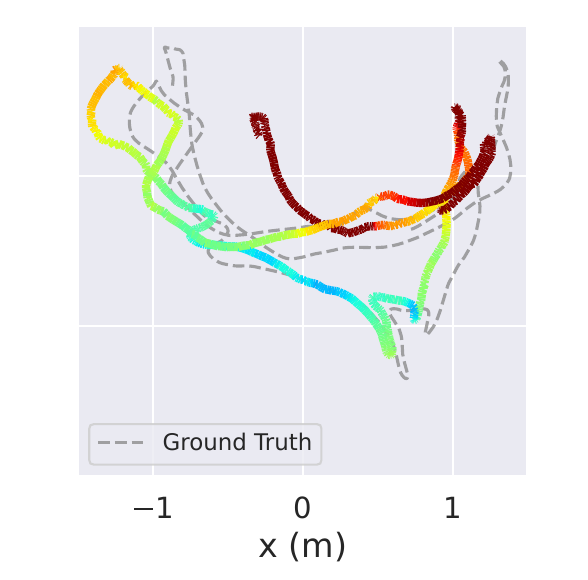} &
        \includegraphics[valign=c, height=\plotheighttop, trim={1.4cm 1cm -0.3cm 0cm}, clip]{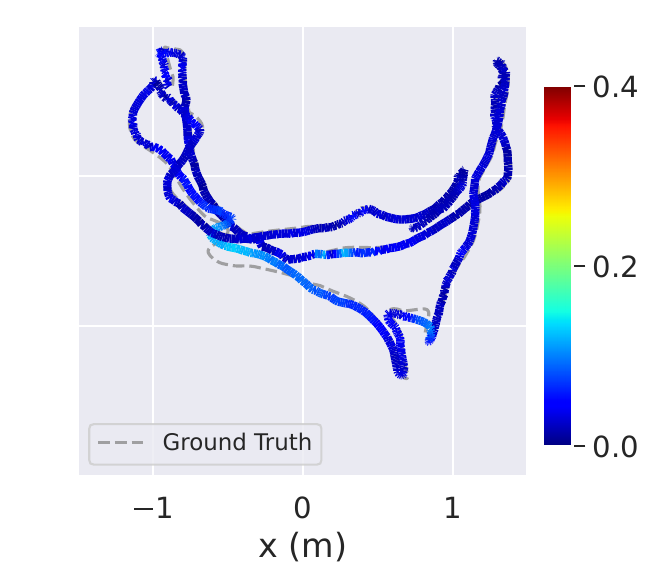} \\[-1pt] 
        
        \raisebox{0.1cm}{\rotatebox{90}{\large\textbf{stairs}}}\hspace{5pt} &
        \includegraphics[valign=c, height=\plotheightbottom, trim={0cm 0 1cm 0cm}, clip]{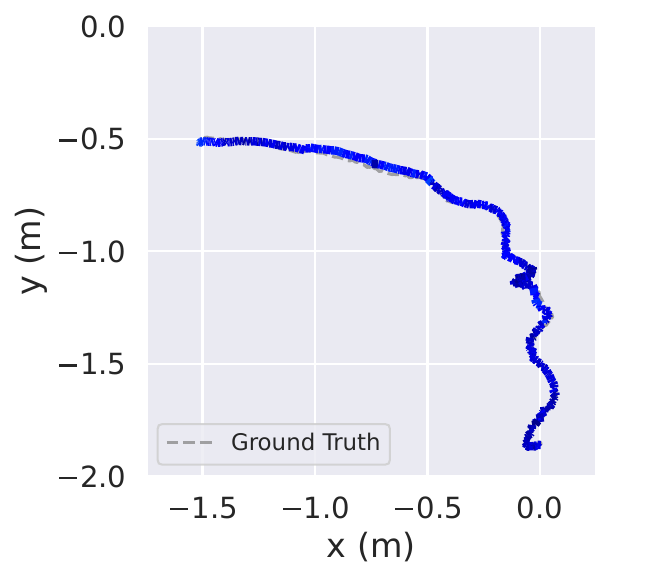} &
        \includegraphics[valign=c, height=\plotheightbottom, trim={1.5cm 0cm 1cm 0cm}, clip]{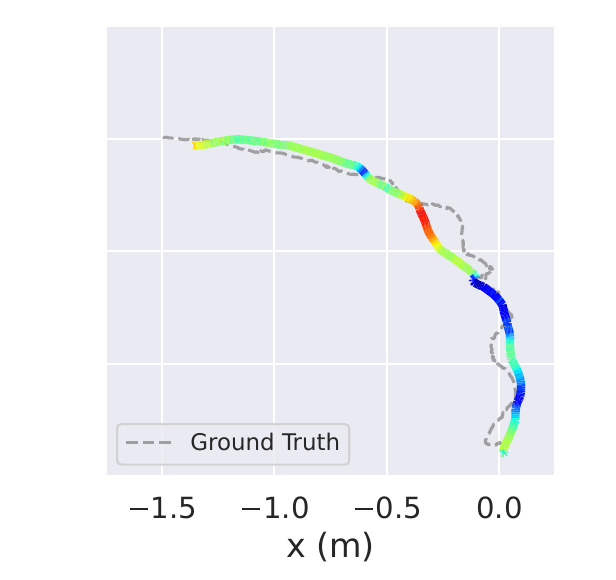} &
        \includegraphics[valign=c, height=\plotheightbottom, trim={1.5cm 0 1cm 0cm}, clip]{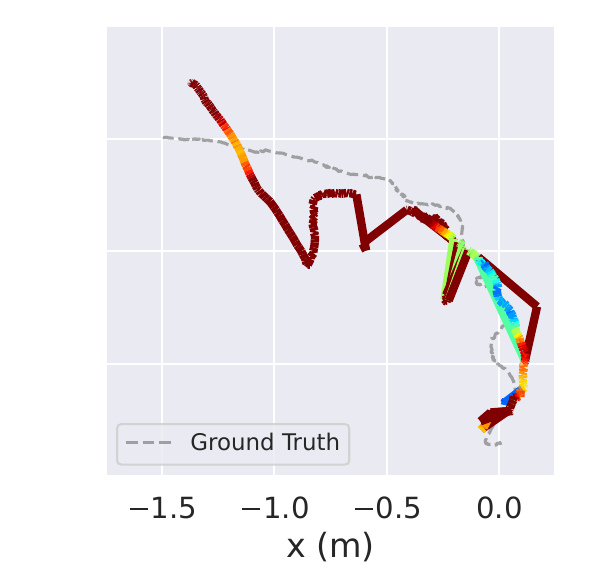} &
        \includegraphics[valign=c, height=\plotheightbottom, trim={1.5cm 0 -0.3cm 0cm}, clip]{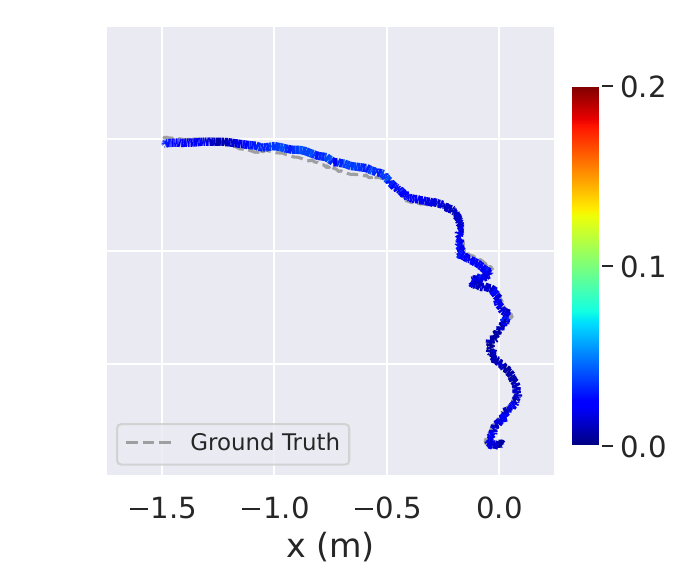} \\
    \end{tabular}%
    } 
    \caption{\textbf{Trajectory estimation results on TUM-RGBD~\cite{sturm2012ate} (\texttt{room}) and 7-Scenes~\cite{shotton20137scenes} (\texttt{stairs}).} 
    Trajectory color encodes the ATE RMSE. Our method demonstrates robust tracking comparable to supervised methods like DroidCalib~\cite{hagemann2023droidcalib}, whereas the photometric-only baseline struggles with small parallax and fails significantly in challenging scenes.}
    \label{fig:trajectory_results_tum_7scenes}
\end{figure*}

\subsection{Camera trajectory evaluation}

\boldparagraph{Baselines} 
For camera trajectory evaluation, we compare \Ourname against three categories of methods: learning-based approaches whose underlying models are fully supervised (DROID-SLAM~\cite{teed2021DROID}, DroidCalib~\cite{hagemann2023droidcalib}, TartanVO~\cite{corl2020tartanvo}, DeepV2D~\cite{teed2020deepv2d}, MASt3R-SLAM~\cite{murai2025mast3rslam}), classical geometric methods (DSO~\cite{engel2017dso}, ORB-SLAM2/3~\cite{mur2017orbslam2,campos2021orbslam3}), and self-supervised methods (D$^3$FlowSLAM~\cite{yu2022d3flowslam}, Dynamo-Depth~\cite{sun2023dynamo}, AnyCam~\cite{wimbauer2025anycam}). 

\boldparagraph{Results} 
We report ATE RMSE in
\Cref{tab:tartanair_results,tab:tum_results,tab:euroc_results}. \Ourname
establishes a new state of the art among self-supervised methods, outperforming
D$^3$FlowSLAM~\cite{yu2022d3flowslam} by $37\%$, $68\%$, and $96\%$ on TartanAir,
TUM-RGBD, and EuRoC, respectively. Notably, D$^3$FlowSLAM requires fine-tuning
on evaluation-domain sequences to prevent divergence, whereas our model achieves
these results with fixed pre-trained weights and no test-time adaptation.
Compared with fully supervised systems, \Ourname performs comparably to
DroidCalib~\cite{hagemann2023droidcalib} and DROID-SLAM~\cite{teed2021DROID} on
EuRoC and TUM-RGBD, while consistently outperforming earlier baselines such as
DeepV2D~\cite{teed2020deepv2d} and TartanVO~\cite{corl2020tartanvo}. Moreover,
\textit{Ours (SIFT + NN)} performs close to the LightGlue-based \textit{Ours}
while still surpassing all prior self-supervised baselines, indicating that our
formulation is robust to the choice of feature matcher.

Across all benchmarks, our method consistently outperforms the photometric-only
baseline, which achieves reasonable accuracy in simple scenes but remains
susceptible to catastrophic tracking failures. Sparse correspondence guidance
stabilizes optimization and reduces drift, yielding smooth trajectories closely
aligned with ground truth (see \Cref{fig:trajectory_results_tum_7scenes}).

\subsection{Dense reconstruction evaluation}

\boldparagraph{Baselines}
For dense reconstruction, we compare against AnyCam~\cite{wimbauer2025anycam}, DroidCalib~\cite{hagemann2023droidcalib}, and recent systems built upon feed-forward geometric models, including SLAM3R~\cite{liu2025slam3r}, ViSTA-SLAM~\cite{zhang2025vistaslam}, MASt3R-SLAM~\cite{murai2025mast3rslam}, and VGGT-SLAM~\cite{vggtslam2025} on 7-Scenes~\cite{shotton20137scenes}. D$^3$FlowSLAM is excluded due to the absence of public code. Since AnyCam does not directly produce a dense 3D reconstruction, we follow the original work and use the predicted poses and intrinsics together with UniDepth~\cite{piccinelli2024unidepth} depth predictions.

\boldparagraph{Results}
In \Cref{tab:7scenes}, we evaluate 3D reconstruction quality on
7-Scenes~\cite{shotton20137scenes}. To ensure fair comparison with
intrinsics-free models, we assess our method self-calibrating, i.e. a
shared focal length is jointly optimized during DBA. \Ourname consistently
outperforms the photometric baseline and remains competitive with fully
supervised methods in both trajectory and reconstruction estimation,
demonstrating robustness to calibration uncertainty.

\begin{figure}[t]
    \centering
    \setlength{\tabcolsep}{1pt}
    \renewcommand{\arraystretch}{0.5} 

    \begin{tabular}{cccc}
        &
        \scriptsize \textbf{DroidCalib} &
        \scriptsize \textbf{photo. only} &
        \scriptsize \textbf{Ours} \\
        \noalign{\vspace{2pt}}
        
        \raisebox{-0.5\height}{\rotatebox{90}{\scriptsize \textbf{redkitchen}}} &
        \raisebox{-0.5\height}{\includegraphics[width=0.3\linewidth, trim={0cm 0cm 0cm 0cm}, clip]{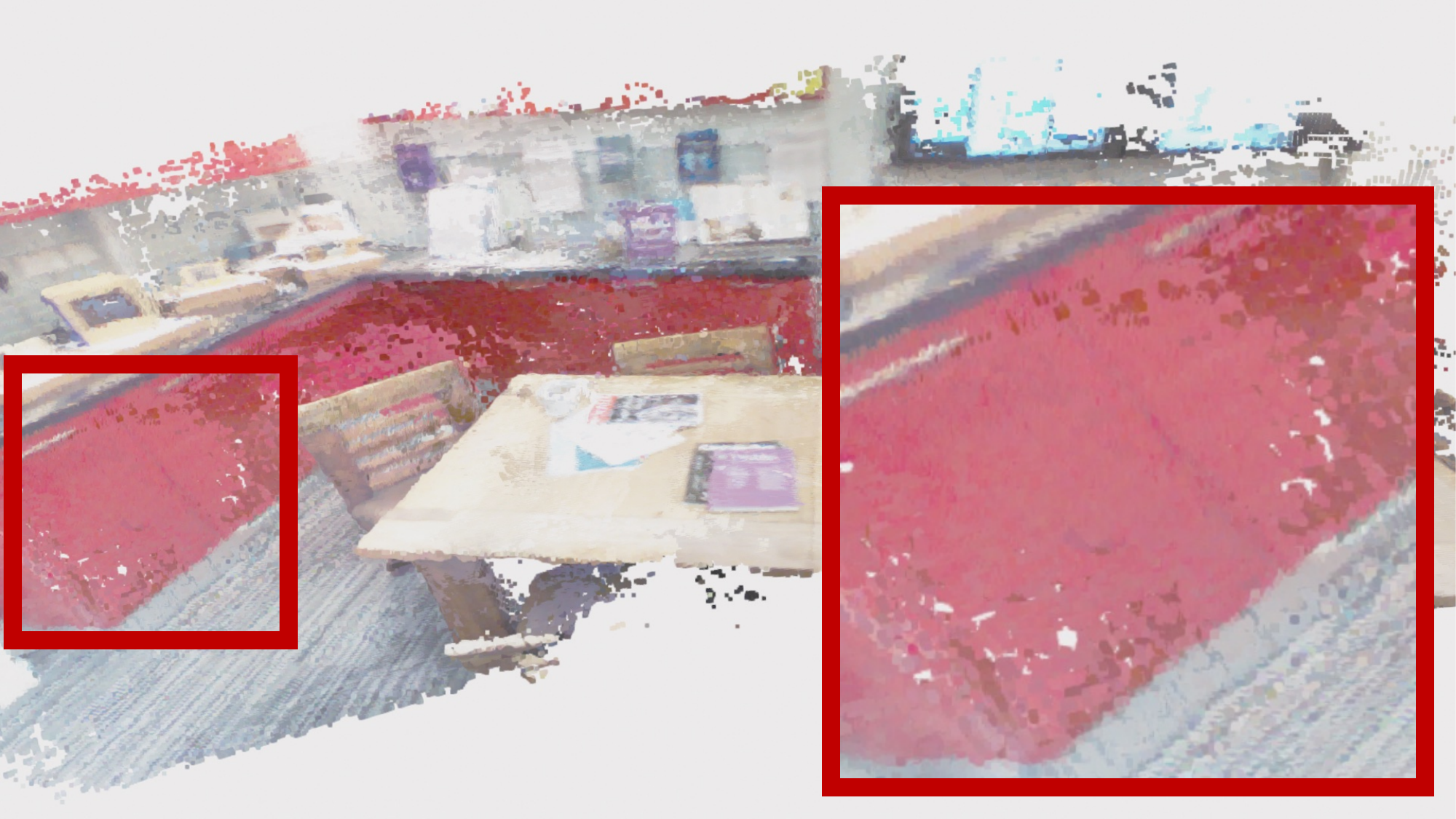}} &
        \raisebox{-0.5\height}{\includegraphics[width=0.3\linewidth, trim={0cm 0cm 0cm 0cm}, clip]{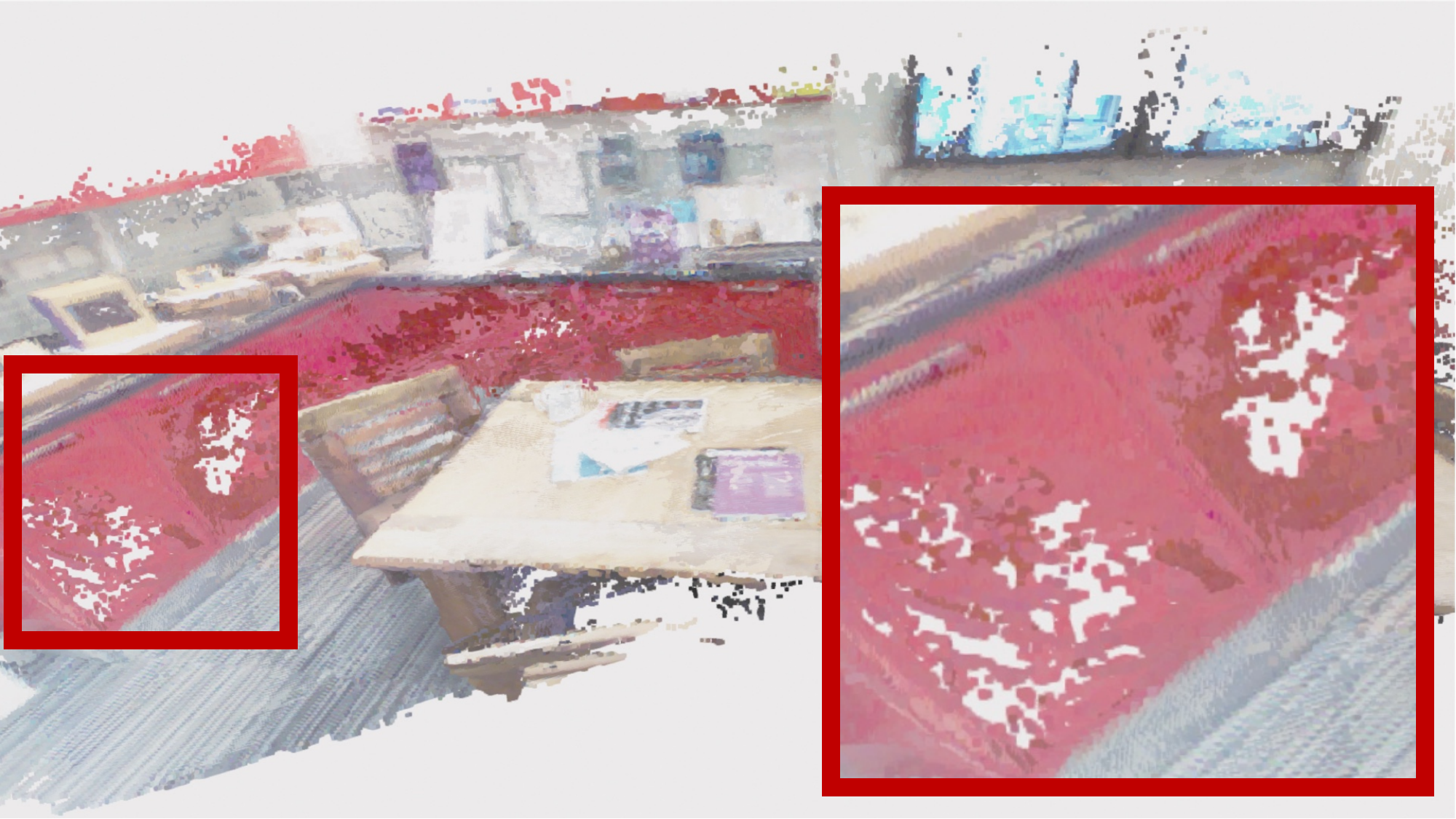}} &
        \raisebox{-0.5\height}{\includegraphics[width=0.3\linewidth, trim={0cm 0cm 0cm 0cm}, clip]{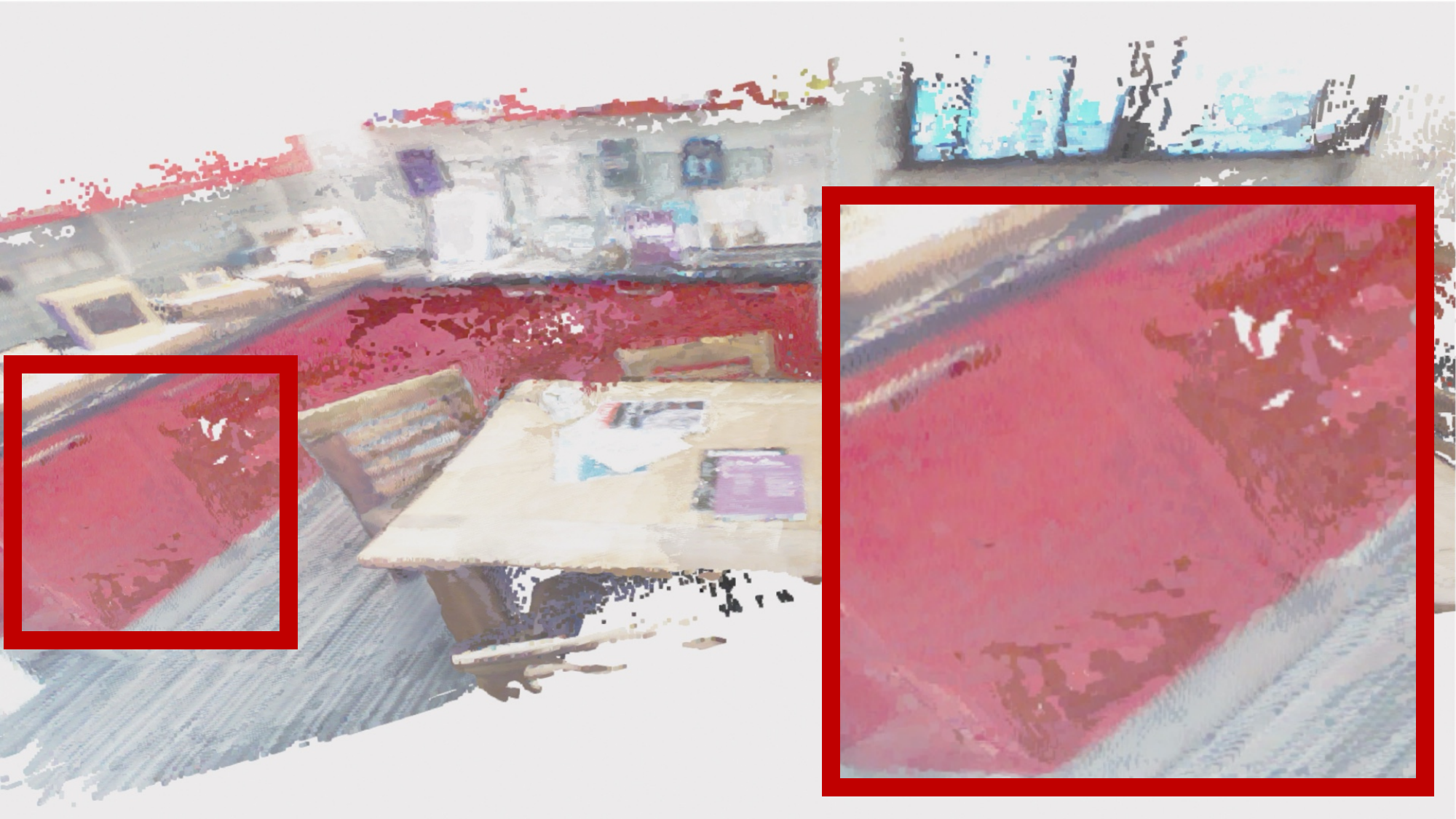}} \\
        \noalign{\vspace{2pt}}
        
        \raisebox{-0.5\height}{\rotatebox{90}{\scriptsize \textbf{room}}} &
        \raisebox{-0.5\height}{\includegraphics[width=0.3\linewidth, trim={0cm 0cm 0cm 0cm}, clip]{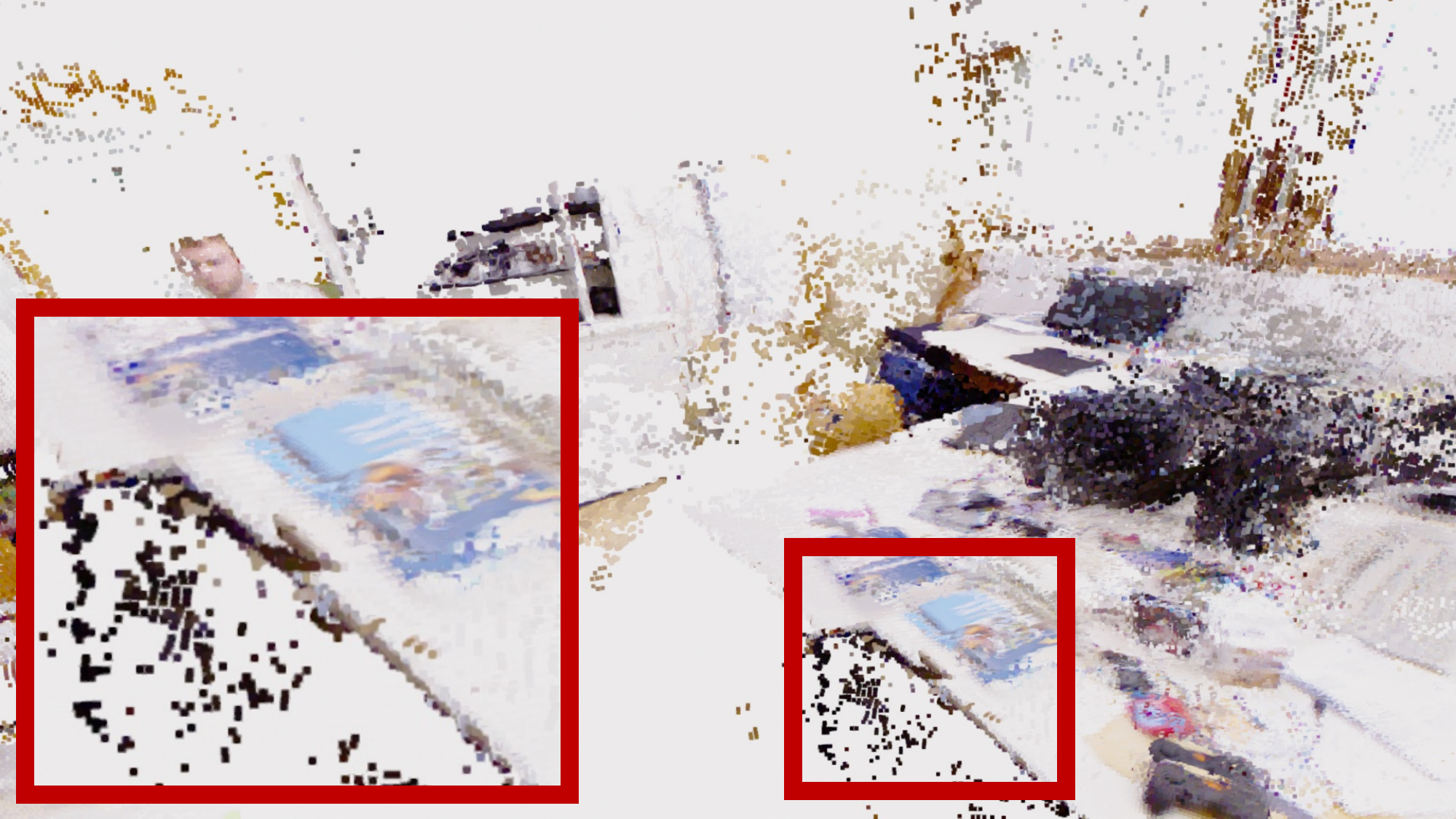}} &
        \raisebox{-0.5\height}{\includegraphics[width=0.3\linewidth, trim={0cm 0cm 0cm 0cm}, clip]{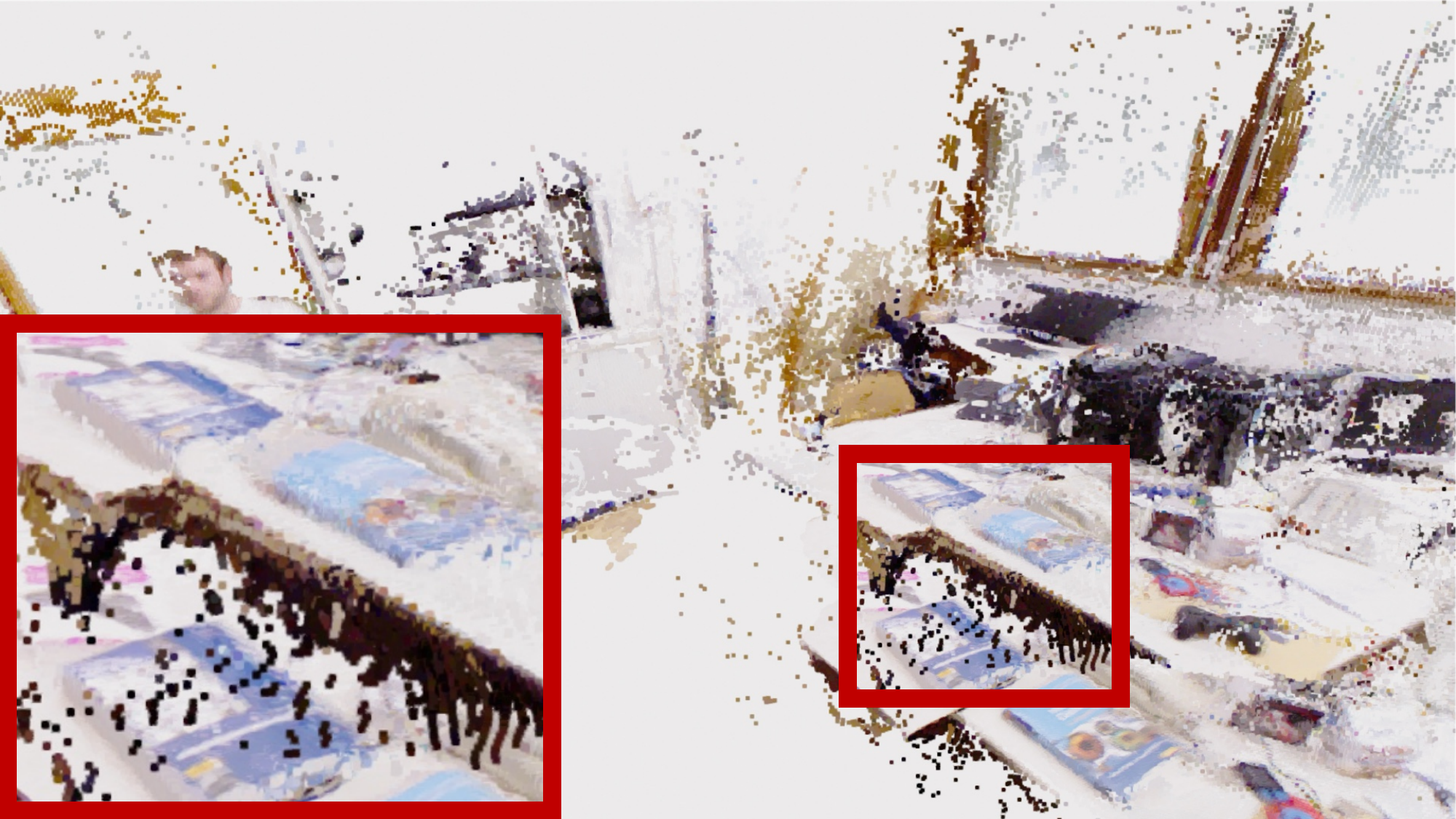}} &
        \raisebox{-0.5\height}{\includegraphics[width=0.3\linewidth, trim={0cm 0cm 0cm 0cm}, clip]{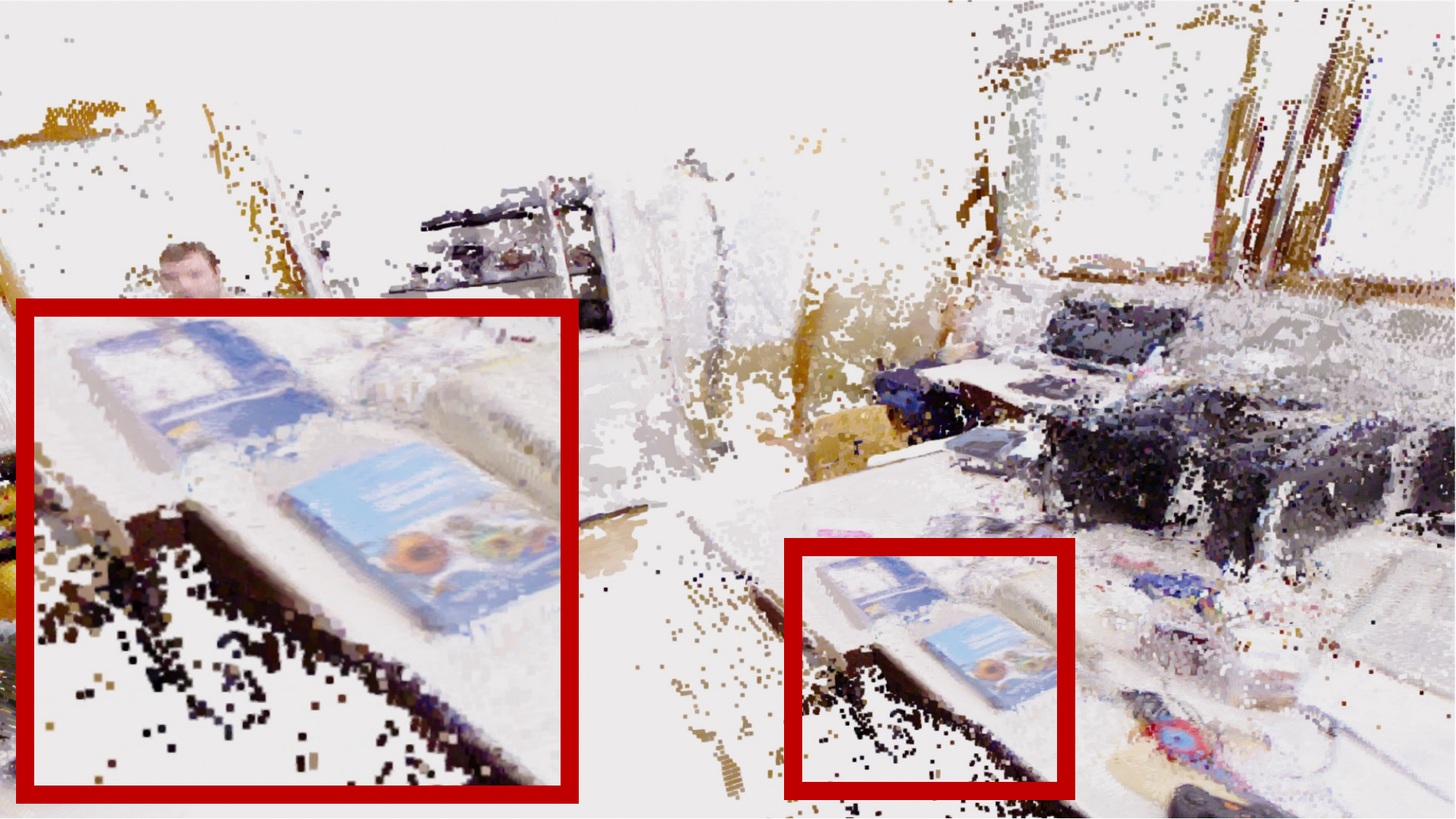}} \\
        \noalign{\vspace{2pt}}

        \raisebox{-0.5\height}{\rotatebox{90}{\scriptsize \textbf{0054\_00}}} &
        \raisebox{-0.5\height}{\includegraphics[width=0.3\linewidth, trim={0cm 0cm 0cm 0cm}, clip]{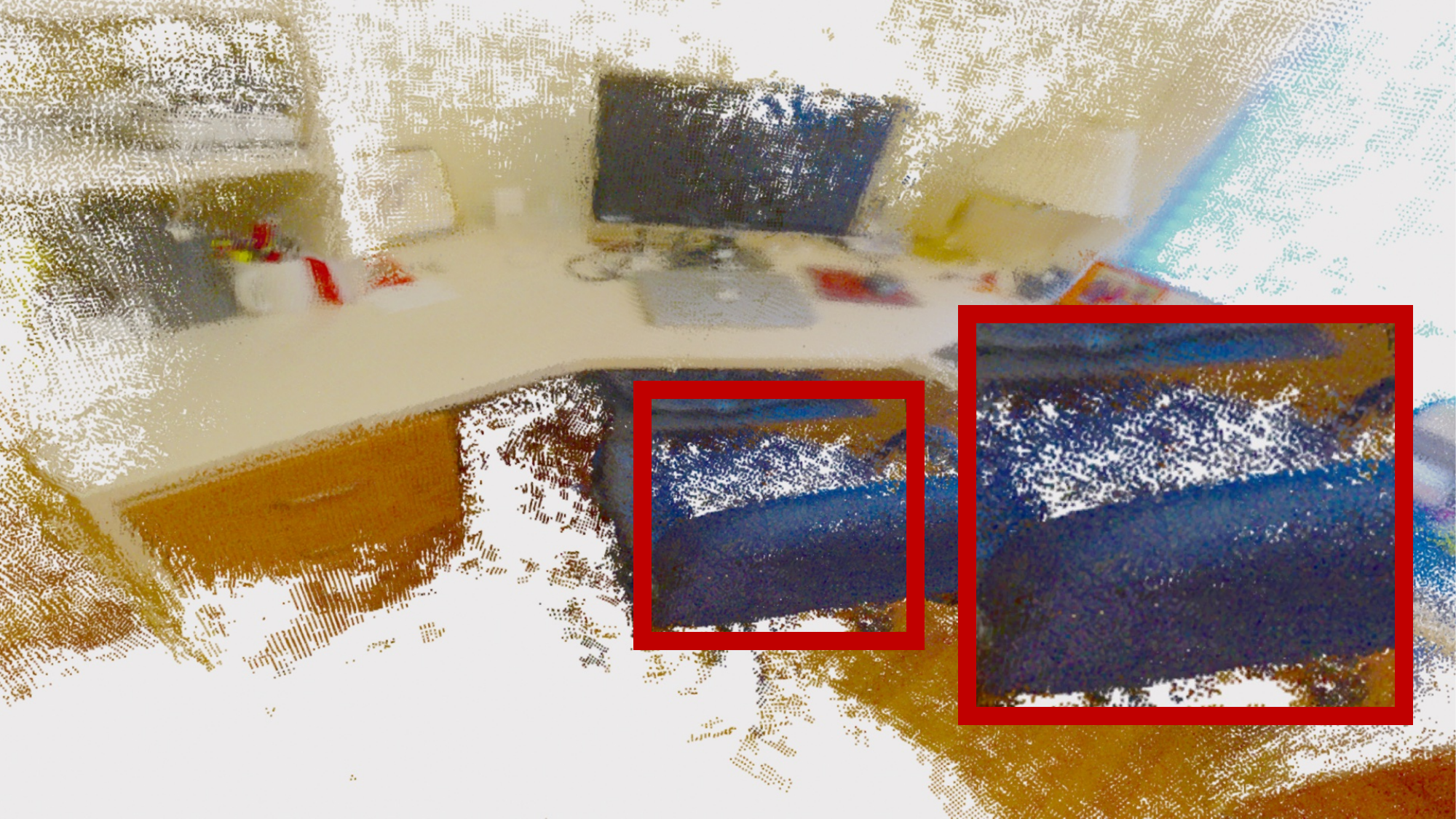}} &
        \raisebox{-0.5\height}{\includegraphics[width=0.3\linewidth, trim={0cm 0cm 0cm 0cm}, clip]{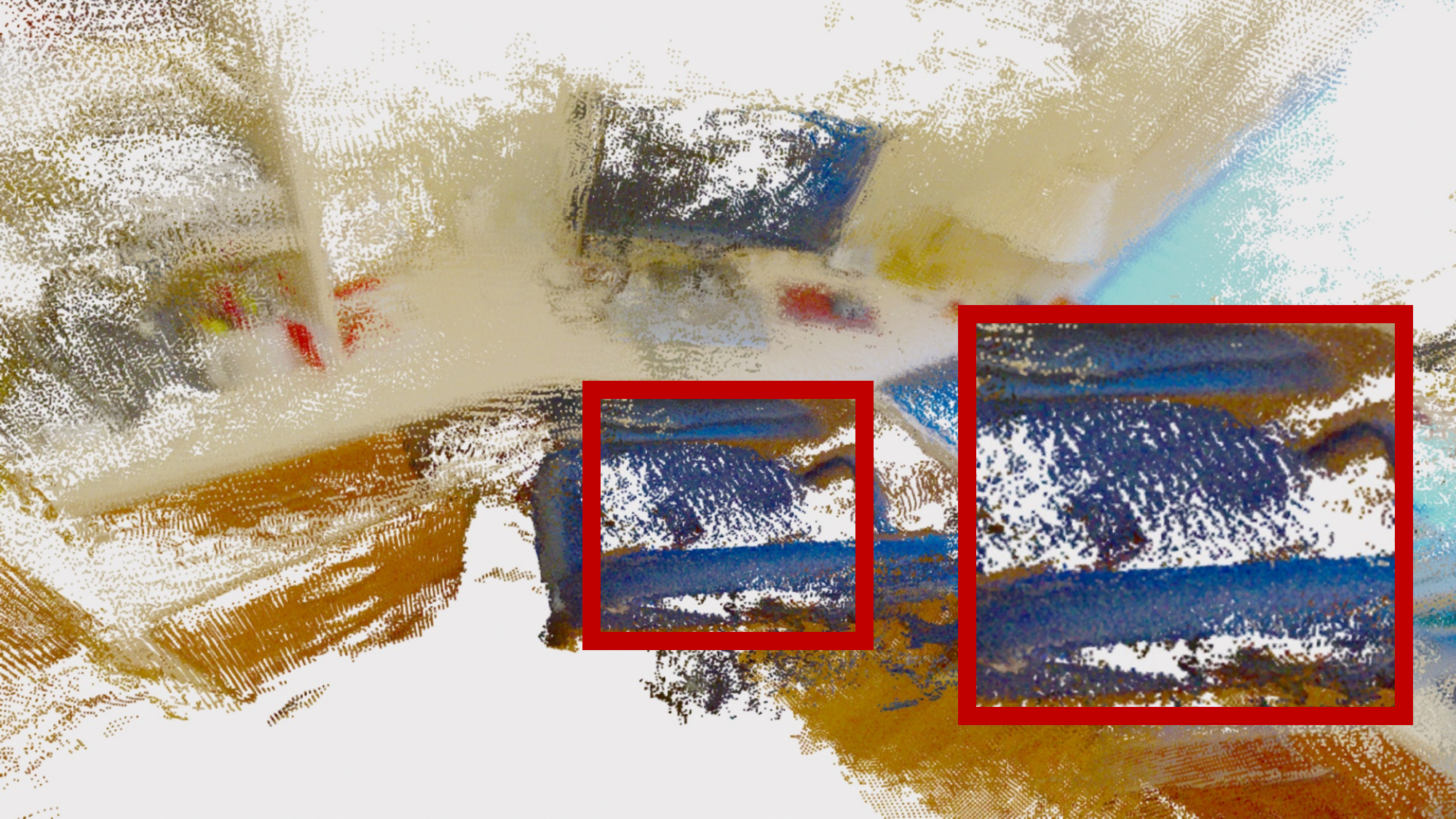}} &
        \raisebox{-0.5\height}{\includegraphics[width=0.3\linewidth, trim={0cm 0cm 0cm 0cm}, clip]{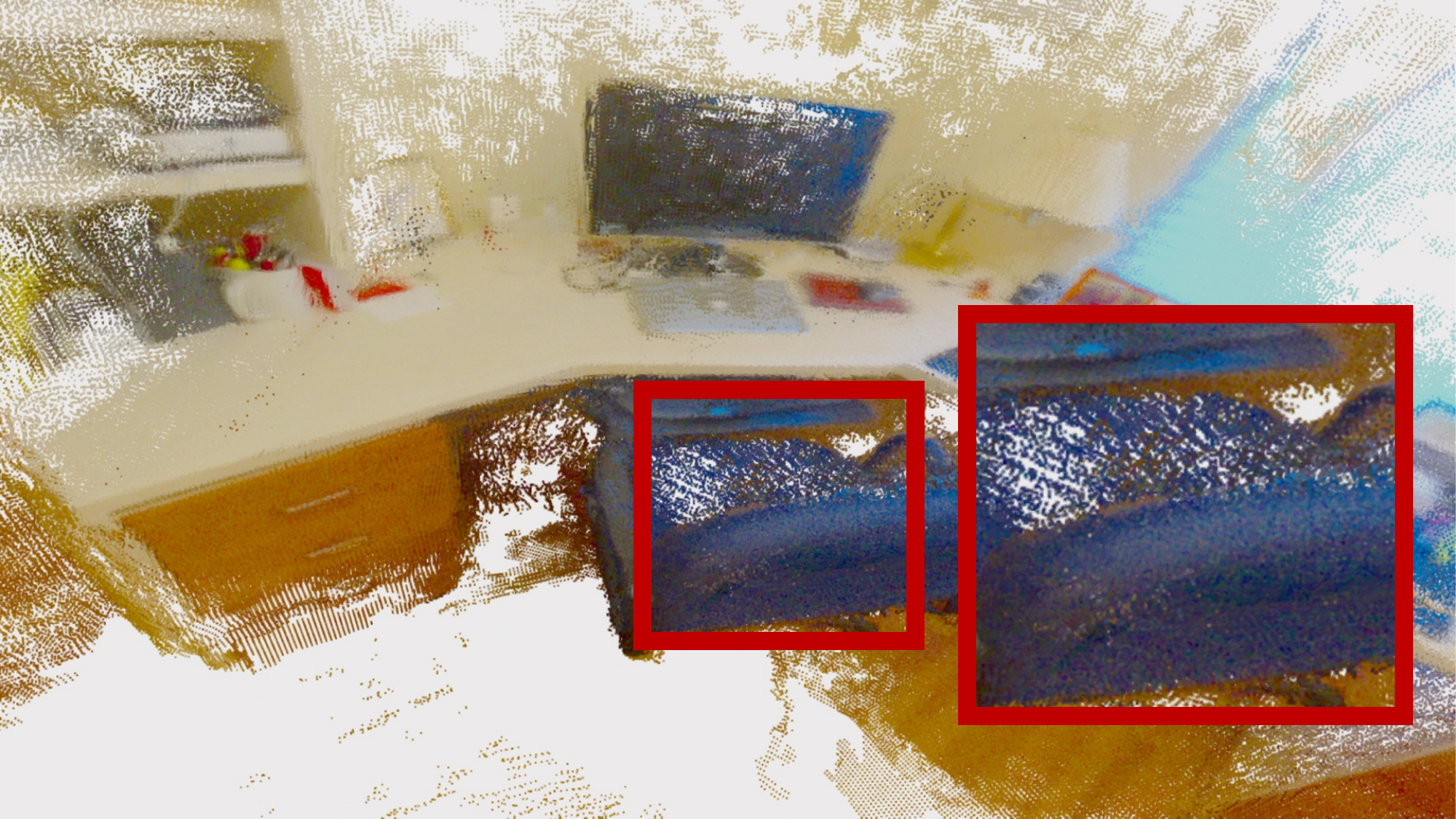}} \\
    \end{tabular}
    \caption{\textbf{Reconstruction results on 7-Scenes~\cite{shotton20137scenes} (\texttt{redkitchen}), TUM-RGBD~\cite{sturm2012ate} (\texttt{room}) and ScanNet~\cite{dai2017scannet} (\texttt{0054\_00}).} Red boxes highlight regions where the photometric-only baseline exhibits misalignment, while \Ourname matches the supervised DroidCalib~\cite{hagemann2023droidcalib}.}     
    \label{fig:recon}
\end{figure}

\Cref{fig:recon} presents qualitative 3D reconstructions for
DroidCalib~\cite{hagemann2023droidcalib}, \Ourname and the photometric-only
baseline. On the relatively simple 7-Scenes
\texttt{redkitchen}~\cite{shotton20137scenes} sequence, all methods perform
comparably, though the photometric baseline produces slightly sparser geometry.
In the more challenging TUM-RGBD \texttt{room}~\cite{sturm2012ate} and ScanNet
\texttt{0054\_00}~\cite{dai2017scannet} scene, \Ourname remains closely aligned
with ground truth and competitive with DroidCalib, whereas the photometric
baseline exhibits noticeable misalignment.

\begin{table}[t]
\centering
\caption{
\textbf{Tracking and Reconstruction Evaluation on 7-Scenes~\cite{shotton20137scenes}.}
\Ourname achieves competitive performance with fully supervised methods.
}
\label{tab:7scenes}
\resizebox{\columnwidth}{!}{%
\begin{tabular}{clcccc}
    \toprule
    \multicolumn{2}{c}{\multirow{2}{*}{\textbf{Method}}}
    & \multicolumn{1}{c}{\textbf{Trajectory}}
    & \multicolumn{3}{c}{\textbf{Reconstruction}} \\
    \cmidrule(lr){3-3}
    \cmidrule(lr){4-6}

    \multicolumn{2}{c}{}
    & ATE
    & Acc.
    & Comp.
    & Cham. \\
    \midrule

    \multirow{5}{*}{\rotatebox[origin=c]{90}{\color{gray}\textit{Sup.}}}
    & \color{gray}SLAM3R~\cite{liu2025slam3r}
    & \color{gray}0.098
    & \color{gray}0.053
    & \color{gray}0.059
    & \color{gray}0.056 \\

    & \color{gray}MASt3R-SLAM~\cite{murai2025mast3rslam}
    & \color{gray}0.066
    & \color{gray}0.059
    & \color{gray}\textbf{0.056}
    & \color{gray}0.057 \\

    & \color{gray}VGGT-SLAM~\cite{vggtslam2025}
    & \color{gray}0.070
    & \color{gray}0.052
    & \color{gray}0.061
    & \color{gray}0.056 \\

    & \color{gray}ViSTA-SLAM~\cite{zhang2025vistaslam}
    & \color{gray}0.055
    & \color{gray}0.045
    & \color{gray}\textbf{0.056}
    & \color{gray}0.051 \\

    & \color{gray}DroidCalib~\cite{hagemann2023droidcalib}
    & \color{gray}\textbf{0.053}
    & \color{gray}\textbf{0.024}
    & \color{gray}0.068
    & \color{gray}\textbf{0.046} \\
    \midrule

    \multirow{4}{*}{\rotatebox[origin=c]{90}{\textit{S.-S.}}}
    & AnyCam
    & 0.207
    & 0.250
    & 0.141
    & 0.195 \\

    & Ours (photo. only)
    & \cellcolor{yellow!25}0.086
    & \cellcolor{yellow!25}0.082
    & \cellcolor{orange!25}0.084
    & \cellcolor{yellow!25}0.083 \\

    & Ours (SIFT + NN)
    & \cellcolor{orange!25}0.082
    & \cellcolor{orange!25}0.069
    & \cellcolor{yellow!25}0.092
    & \cellcolor{orange!25}0.080 \\

    & \textbf{Ours}
    & \cellcolor{red!25}\textbf{0.061}
    & \cellcolor{red!25}\textbf{0.026}
    & \cellcolor{red!25}\textbf{0.065}
    & \cellcolor{red!25}\textbf{0.045} \\
    \bottomrule
\end{tabular}%
}

\vspace{-1em}
\end{table}

\subsection{Ablation study} \label{sec:ablation}

To validate our self-supervised loss formulation, we conduct an ablation study
restricted to the TartanAir \texttt{Hospital} environment to keep the training
overhead manageable. We train on \texttt{easy P000-P028} and \texttt{hard
P037-P046}, and evaluate on \texttt{P029-P036} and \texttt{P047-P049}
(see~\Cref{tab:ablation}).

\boldparagraph{Loss components}
We first analyze the contribution of individual loss components in
\Cref{tab:ablation}. Sparse correspondence supervision alone performs poorly,
indicating that sparse geometric constraints are insufficient to guide dense
optimization. Photometric supervision provides a substantially stronger dense
baseline, while combining it with our sparse correspondence loss further improves
all metrics by a large margin. These results demonstrate the complementary roles
of dense photometric guidance and sparse geometric constraints.

\begin{table}[t]
    \centering
    \caption{
        \textbf{Ablation study.}
        Quantitative evaluation on the TartanAir~\cite{wang2020tartanair}
        \texttt{Hospital} environment.
    }
    \label{tab:ablation}

    \resizebox{\columnwidth}{!}{%
    \begin{tabular}{clccccc}
        \toprule
        \multicolumn{2}{c}{\multirow{2}{*}{\textbf{Supervision}}}
        & \multicolumn{1}{c}{\textbf{Trajectory}}
        & \multicolumn{3}{c}{\textbf{Reconstruction}}
        & \multicolumn{1}{c}{\textbf{Flow}} \\
        \cmidrule(lr){3-3}
        \cmidrule(lr){4-6}
        \cmidrule(lr){7-7}

        \multicolumn{2}{c}{}
        & ATE [m]
        & Acc.
        & Comp.
        & Cham.
        & EPE \\
        \midrule

        \multirow{3}{*}{\rotatebox[origin=c]{90}{\textit{Loss}}}
        & photo. only
        & 0.281 & 0.079 & 0.135 & 0.107 & 2.389 \\

        & corres. only
        & 0.633 & 0.245 & 0.235 & 0.240 & 1.920 \\

        & \textbf{Ours}
        & \textbf{0.048}
        & \textbf{0.031}
        & \textbf{0.090}
        & \textbf{0.061}
        & \textbf{1.580} \\

        \midrule

        \multirow{4}{*}{\rotatebox[origin=c]{90}{\textit{Sparsity}}}
        & GT corres. ($196k$)
        & 0.017 & 0.027 & 0.077 & 0.052 & 0.844 \\

        & GT corres. ($20k$)
        & 0.020 & 0.027 & 0.081 & 0.056 & 0.900 \\

        & GT corres. ($2k$)
        & 0.018 & 0.028 & 0.079 & 0.053 & 0.850 \\

        & Ours ($<2k$)
        & 0.048 & 0.031 & 0.090 & 0.061 & 1.580 \\

        \midrule

        \multirow{3}{*}{\rotatebox[origin=c]{90}{\textit{Corres.}}}
        & Uniform sampling + GT
        & 0.019 & 0.025 & 0.075 & 0.050 & 0.890 \\

        & SuperPoint + GT
        & 0.027 & 0.026 & 0.083 & 0.055 & 1.133 \\

        & SuperPoint + LightGlue
        & 0.048 & 0.031 & 0.090 & 0.061 & 1.580 \\

        \bottomrule
    \end{tabular}%
    }

    \vspace{-1em}
\end{table}

\boldparagraph{Correspondence sparsity}
To examine whether match sparsity limits performance, we replace our generated
correspondences with uniformly sampled ground-truth (GT) correspondences. As
shown in \Cref{tab:ablation}, using only $2k$ sparse GT correspondences achieves
performance nearly identical to $196k$ dense GT correspondences, indicating that
dense supervision is not inherently required when stabilized by our continuous
photometric loss. This suggests that the remaining gap between our method and
the GT setting is driven primarily by the quality and distribution of the
available matches rather than by their quantity.

\boldparagraph{Correspondence quality}
We disentangle the effects of correspondence noise and non-uniform spatial
coverage by evaluating three settings with the same number of correspondences on
the TartanAir~\cite{wang2020tartanair} \texttt{Hospital} split: uniformly
sampled GT correspondences, GT correspondences at
SuperPoint~\cite{detone2018superpoint} locations, and
SuperPoint--LightGlue~\cite{lindenberger2023lightglue} matches. As shown in
\Cref{tab:ablation}, restricting GT correspondences to SuperPoint locations
degrades performance, revealing the effect of non-uniform coverage, while
replacing GT targets with LightGlue matches causes a further drop due to
correspondence noise.

\subsection{Self-supervised adaptation and data scalability} \label{sec:adaptation}

\boldparagraph{Domain adaptation}
We test whether self-supervised training on unlabeled target-domain data
can adapt a pretrained model and improve generalization to unseen sequences. We
continue training \Ourname on ScanNet~\cite{dai2017scannet} scenes
\texttt{0000--0049} using the same self-supervised losses and no ground-truth
pose or depth labels, then evaluate on the disjoint held-out scenes
\texttt{0100--0107}. As shown in \Cref{tab:scannet_finetune}, the mean ATE
decreases from 0.309\,m to 0.166\,m, a relative reduction of 47\%, with
improvements on seven of eight scenes.

\begin{table}[t]
    \centering
    \caption{
        \textbf{Domain adaptation on ScanNet}~\cite{dai2017scannet},
        evaluated on held-out scenes \texttt{0100--0107}: ATE RMSE [m].
    }
    \label{tab:scannet_finetune}

    \setlength{\tabcolsep}{2pt}

    \resizebox{\columnwidth}{!}{%
    \begin{tabular}{lccccccccc}
        \toprule
        \textbf{Method}
        & \texttt{100}
        & \texttt{101}
        & \texttt{102}
        & \texttt{103}
        & \texttt{104}
        & \texttt{105}
        & \texttt{106}
        & \texttt{107}
        & \textbf{Avg.} \\
        \midrule

        Ours
        & 0.034
        & 0.871
        & \textbf{0.096}
        & 0.562
        & 0.289
        & 0.151
        & 0.403
        & 0.069
        & 0.309 \\

        Ours (adapt.)
        & \textbf{0.033}
        & \textbf{0.733}
        & 0.098
        & \textbf{0.053}
        & \textbf{0.098}
        & \textbf{0.128}
        & \textbf{0.115}
        & \textbf{0.068}
        & \textbf{0.166} \\

        \bottomrule
    \end{tabular}%
   }

\end{table}

\boldparagraph{Data scalability}
We further evaluate whether the same self-supervised objective can benefit from
additional unlabeled data outside the evaluation domain. We continue training
\Ourname on DL3DV~\cite{ling2024dl3dv}, a large-scale collection of real-world
videos without the dense geometric ground truth required for conventional
supervised training. We then evaluate on the disjoint
TartanAir~\cite{wang2020tartanair} hard split, whose data distribution differs
from the DL3DV training domain. As shown in
\Cref{tab:tartanair_results__finetune}, the mean ATE decreases from 0.50\,m to
0.37\,m, showing that additional real-world training data can improve
cross-domain generalization.

\noindent These results show that our self-supervised objective not only enables domain adaptation but also scales with abundant unlabeled real-world video, offering a practical path to continued improvement without costly ground-truth annotation.

\begin{table}[t]
    \centering
    \caption{
        \textbf{Continued self-supervised training on DL3DV}~\cite{ling2024dl3dv},
        evaluated on the TartanAir~\cite{wang2020tartanair} hard split:
        ATE RMSE [m]. Sequence labels omit the \texttt{MH} prefix.
    }
    \label{tab:tartanair_results__finetune}

    \setlength{\tabcolsep}{2pt}

    \resizebox{\columnwidth}{!}{%
    \begin{tabular}{lccccccccc}
        \toprule
        \textbf{Method}
        & \texttt{000}
        & \texttt{001}
        & \texttt{002}
        & \texttt{003}
        & \texttt{004}
        & \texttt{005}
        & \texttt{006}
        & \texttt{007}
        & \textbf{Avg.} \\
        \midrule

        Ours
        & 0.36
        & \textbf{0.04}
        & 0.03
        & 0.05
        & 2.04
        & 0.88
        & \textbf{0.38}
        & 0.24
        & 0.50 \\

        Ours + DL3DV
        & \textbf{0.29}
        & 0.07
        & \textbf{0.02}
        & \textbf{0.04}
        & \textbf{1.54}
        & \textbf{0.20}
        & 0.68
        & \textbf{0.12}
        & \textbf{0.37} \\

        \bottomrule
    \end{tabular}%
    }
\vspace{-1em}
\end{table}

\section{Conclusion}

We present \Ourname, a self-supervised visual SLAM framework that integrates
sparse correspondence guidance with dense photometric constraints, eliminating
the need for ground-truth pose and depth annotations during training. Extensive
experiments show that incorporating sparse correspondences significantly
improves both trajectory estimation and 3D reconstruction, enabling our method
to outperform existing self-supervised approaches and achieve performance close
to fully supervised baselines. Future work could extend our framework to modern
feed-forward reconstruction models and relax the current assumption of static
scenes, enabling training and inference on videos with dynamic contents.

\bibliographystyle{IEEEtran}
\bibliography{main}

\clearpage

\let\savedRASection\section
\appendix
\let\section\savedRASection

\setcounter{section}{0}
\renewcommand{\thesection}{\Alph{section}}
\makeatletter
\@ifpackageloaded{hyperref}{\renewcommand{\theHsection}{supp.\Alph{section}}}{}
\makeatother

This supplementary material provides \textbf{additional results} in
\Cref{sec:supp_additional_experiments}, including qualitative and quantitative
comparisons. We also provide more details regarding the \textbf{technical
implementation} in \Cref{sec:supp_implementation_details} and discuss
\textbf{limitations and further ideas for future work} in
\Cref{sec:supp_limit_future}.

\section{Additional experiments}
\label{sec:supp_additional_experiments}

\subsection{Dense reconstruction}

\begin{table*}[t]
\centering
\caption{\textbf{Tracking and reconstruction results on Replica~\cite{straub2019replica}.} \Ourname performs on par with supervised baselines.}
\vspace{0.5em}
\label{tab:replica_detailed}

\renewcommand{\arraystretch}{0.9}

\resizebox{0.9\textwidth}{!}{%
\setlength{\aboverulesep}{1pt}
\setlength{\belowrulesep}{1pt}
\footnotesize
\setlength{\tabcolsep}{5pt}

\begin{tabular}{llccccccccc}
\toprule

\textbf{Method}
& \textbf{Metric}
& \texttt{room0}
& \texttt{room1}
& \texttt{room2}
& \texttt{office0}
& \texttt{office1}
& \texttt{office2}
& \texttt{office3}
& \texttt{office4}
& \textbf{Avg.} \\

\midrule

\multirow{4}{*}{DROID-SLAM~\cite{teed2021DROID}}
& Accuracy
& 0.030 & 0.023 & 0.019 & 0.016 & 0.017 & 0.023 & 0.025 & 0.024 & 0.022 \\
& Completeness
& 0.040 & 0.025 & 0.042 & 0.022 & 0.028 & 0.037 & 0.049 & 0.041 & 0.035 \\
& Chamfer
& 0.035 & 0.024 & 0.031 & 0.020 & 0.022 & 0.030 & 0.037 & 0.033 & 0.029 \\
& ATE
& 0.009 & 0.005 & 0.004 & 0.004 & 0.005 & 0.007 & 0.008 & 0.009 & \textbf{0.006} \\

\midrule

\multirow{4}{*}{DroidCalib~\cite{hagemann2023droidcalib}}
& Accuracy
& 0.034 & 0.021 & 0.019 & 0.017 & 0.015 & 0.022 & 0.027 & 0.025 & 0.023 \\
& Completeness
& 0.042 & 0.026 & 0.040 & 0.024 & 0.027 & 0.038 & 0.047 & 0.043 & 0.036 \\
& Chamfer
& 0.038 & 0.024 & 0.029 & 0.021 & 0.021 & 0.030 & 0.037 & 0.034 & 0.029 \\
& ATE
& 0.010 & 0.007 & 0.006 & 0.006 & 0.007 & 0.008 & 0.010 & 0.008 & 0.008 \\

\midrule

\multirow{4}{*}{Ours}
& Accuracy
& 0.022 & 0.017 & 0.017 & 0.013 & 0.017 & 0.018 & 0.020 & 0.016 & \textbf{0.018} \\
& Completeness
& 0.036 & 0.020 & 0.038 & 0.018 & 0.029 & 0.033 & 0.037 & 0.034 & \textbf{0.031} \\
& Chamfer
& 0.029 & 0.018 & 0.027 & 0.016 & 0.022 & 0.026 & 0.029 & 0.025 & \textbf{0.024} \\
& ATE
& 0.008 & 0.005 & 0.004 & 0.006 & 0.005 & 0.008 & 0.008 & 0.006 & \textbf{0.006} \\

\bottomrule
\end{tabular}%
}
\end{table*}

\boldparagraph{Replica} \label{sec:supp__replica}
In \Cref{tab:replica_detailed}, we present tracking and reconstruction results
on the Replica~\cite{straub2019replica} dataset. As on the
TUM-RGBD~\cite{sturm2012ate} dataset, \Ourname shows strong performance across
all scenes rivaling fully supervised DROID-SLAM~\cite{teed2021DROID} and
DroidCalib~\cite{hagemann2023droidcalib}. Notably, our method yields high
geometric precision, achieving an average accuracy of 0.018 and a chamfer
distance of 0.024 across the evaluated scenes. This demonstrates the robust
generalization capabilities of \Ourname, effectively reconstructing unseen
complex indoor environments in a strictly zero-shot setting.

\boldparagraph{7-Scenes} 
Detailed per-scene results for the 7-Scenes dataset~\cite{shotton20137scenes}
are provided in \Cref{tab:7scenes_detailed} to offer deeper diagnostic insights.
Consistent with our observations on other datasets, the elevated average error
across all metrics of the photometric-only baseline is primarily driven by
specific outlier scenes where the model suffers from tracking divergence. In
contrast, our proposed method maintains stability across these challenging
sequences, demonstrating significantly higher tracking robustness.

\begin{table*}[p]
\centering
\caption{\textbf{Per-scene evaluation on the 7-Scenes~\cite{shotton20137scenes} dataset.}}
\vspace{0.5em}
\label{tab:7scenes_detailed}

\renewcommand{\arraystretch}{0.9}

\resizebox{0.9\textwidth}{!}{%
\setlength{\aboverulesep}{1pt}
\setlength{\belowrulesep}{1pt}
\footnotesize
\setlength{\tabcolsep}{5pt}

\begin{tabular}{cllcccccccc}
\toprule

\multicolumn{2}{c}{\textbf{Method}}
& \textbf{Metric}
& \texttt{chess}
& \texttt{fire}
& \texttt{heads}
& \texttt{office}
& \texttt{pump.}
& \texttt{redkit.}
& \texttt{stairs}
& \textbf{Avg.} \\

\midrule

\multirow{36}{*}{\rotatebox[origin=c]{90}{\textit{Uncalibrated}}}
& \multirow{4}{*}{ViSTA-SLAM~\cite{zhang2025vistaslam}}
& Accuracy
& 0.065 & 0.015 & 0.031 & 0.036 & 0.061 & 0.035 & 0.074 & 0.045 \\
& & Completeness
& 0.063 & 0.022 & 0.040 & 0.048 & 0.037 & 0.030 & 0.154 & 0.056 \\
& & Chamfer
& 0.064 & 0.019 & 0.035 & 0.042 & 0.049 & 0.033 & 0.114 & 0.051 \\
& & ATE
& 0.073 & 0.035 & 0.028 & 0.055 & 0.129 & 0.035 & 0.029 & 0.055 \\

\cmidrule(lr){2-11}

& \multirow{4}{*}{SLAM3R~\cite{liu2025slam3r}}
& Accuracy
& 0.043 & 0.022 & 0.020 & 0.035 & 0.072 & 0.062 & 0.116 & 0.053 \\
& & Completeness
& 0.030 & 0.013 & 0.015 & 0.030 & 0.055 & 0.061 & 0.209 & 0.059 \\
& & Chamfer
& 0.037 & 0.018 & 0.017 & 0.033 & 0.064 & 0.061 & 0.162 & 0.056 \\
& & ATE
& 0.089 & 0.048 & 0.036 & 0.088 & 0.196 & 0.102 & 0.126 & 0.098 \\

\cmidrule(lr){2-11}

& \multirow{4}{*}{MASt3R-SLAM~\cite{murai2025mast3rslam}}
& Accuracy
& 0.090 & 0.037 & 0.027 & 0.047 & 0.097 & 0.070 & 0.045 & 0.059 \\
& & Completeness
& 0.055 & 0.024 & 0.021 & 0.053 & 0.054 & 0.036 & 0.149 & 0.056 \\
& & Chamfer
& 0.073 & 0.031 & 0.024 & 0.050 & 0.075 & 0.053 & 0.097 & 0.057 \\
& & ATE
& 0.063 & 0.046 & 0.029 & 0.103 & 0.112 & 0.074 & 0.032 & 0.066 \\

\cmidrule(lr){2-11}

& \multirow{4}{*}{VGGT-SLAM~\cite{vggtslam2025}}
& Accuracy
& 0.029 & 0.014 & 0.031 & 0.041 & 0.128 & 0.036 & 0.087 & 0.052 \\
& & Completeness
& 0.052 & 0.064 & 0.021 & 0.066 & 0.054 & 0.057 & 0.110 & 0.061 \\
& & Chamfer
& 0.040 & 0.039 & 0.026 & 0.054 & 0.091 & 0.047 & 0.098 & 0.056 \\
& & ATE
& 0.037 & 0.026 & 0.022 & 0.103 & 0.147 & 0.063 & 0.095 & 0.070 \\

\cmidrule(lr){2-11}

& \multirow{4}{*}{DroidCalib~\cite{hagemann2023droidcalib}}
& Accuracy
& 0.014 & 0.011 & 0.007 & 0.017 & 0.070 & 0.025 & 0.027 & 0.024 \\
& & Completeness
& 0.055 & 0.029 & 0.025 & 0.069 & 0.119 & 0.046 & 0.134 & 0.068 \\
& & Chamfer
& 0.035 & 0.020 & 0.016 & 0.043 & 0.094 & 0.036 & 0.081 & 0.046 \\
& & ATE
& 0.033 & 0.024 & 0.013 & 0.089 & 0.143 & 0.050 & 0.017 & 0.053 \\

\cmidrule(lr){2-11}

& \multirow{4}{*}{AnyCam~\cite{wimbauer2025anycam}}
& Accuracy
& 0.213 & 0.284 & 0.273 & 0.188 & 0.228 & 0.222 & 0.342 & 0.250 \\
& & Completeness
& 0.068 & 0.100 & 0.033 & 0.099 & 0.083 & 0.090 & 0.512 & 0.141 \\
& & Chamfer
& 0.141 & 0.192 & 0.153 & 0.144 & 0.156 & 0.156 & 0.427 & 0.195 \\
& & ATE
& 0.119 & 0.174 & 0.169 & 0.293 & 0.307 & 0.132 & 0.255 & 0.207 \\

\cmidrule(lr){2-11}

& \multirow{4}{*}{Ours (photo. only)}
& Accuracy
& 0.017 & 0.013 & 0.010 & 0.020 & 0.069 & 0.031 & 0.411 & 0.082 \\
& & Completeness
& 0.056 & 0.035 & 0.029 & 0.060 & 0.097 & 0.044 & 0.264 & 0.084 \\
& & Chamfer
& 0.038 & 0.024 & 0.020 & 0.041 & 0.083 & 0.037 & 0.338 & 0.083 \\
& & ATE
& 0.035 & 0.024 & 0.013 & 0.092 & 0.153 & 0.049 & 0.237 & 0.086 \\

\cmidrule(lr){2-11}

& \multirow{4}{*}{Ours (SIFT + NN)}
& Accuracy
& 0.017 & 0.014 & 0.011 & 0.022 & 0.072 & 0.029 & 0.317 & 0.069 \\
& & Completeness
& 0.061 & 0.037 & 0.026 & 0.067 & 0.095 & 0.045 & 0.311 & 0.092 \\
& & Chamfer
& 0.039 & 0.025 & 0.018 & 0.045 & 0.084 & 0.037 & 0.314 & 0.080 \\
& & ATE
& 0.035 & 0.023 & 0.012 & 0.090 & 0.143 & 0.048 & 0.221 & 0.082 \\

\cmidrule(lr){2-11}

& \multirow{4}{*}{Ours}
& Accuracy
& 0.019 & 0.013 & 0.005 & 0.020 & 0.069 & 0.029 & 0.029 & 0.026 \\
& & Completeness
& 0.058 & 0.035 & 0.026 & 0.063 & 0.097 & 0.046 & 0.127 & 0.065 \\
& & Chamfer
& 0.038 & 0.023 & 0.017 & 0.041 & 0.083 & 0.037 & 0.078 & 0.045 \\
& & ATE
& 0.034 & 0.024 & 0.013 & 0.143 & 0.142 & 0.048 & 0.020 & 0.061 \\

\midrule

\multirow{24}{*}{\rotatebox[origin=c]{90}{\textit{Calibrated}}}
& \multirow{4}{*}{DROID-SLAM~\cite{teed2021DROID}}
& Accuracy
& 0.014 & 0.011 & 0.007 & 0.017 & 0.064 & 0.031 & 0.032 & 0.025 \\
& & Completeness
& 0.056 & 0.030 & 0.029 & 0.060 & 0.126 & 0.052 & 0.161 & 0.073 \\
& & Chamfer
& 0.035 & 0.021 & 0.018 & 0.039 & 0.095 & 0.041 & 0.097 & 0.049 \\
& & ATE
& 0.035 & 0.025 & 0.015 & 0.092 & 0.145 & 0.053 & 0.104 & 0.067 \\

\cmidrule(lr){2-11}

& \multirow{4}{*}{DroidCalib~\cite{hagemann2023droidcalib}}
& Accuracy
& 0.014 & 0.012 & 0.007 & 0.017 & 0.070 & 0.027 & 0.030 & 0.025 \\
& & Completeness
& 0.055 & 0.028 & 0.028 & 0.063 & 0.128 & 0.048 & 0.141 & 0.070 \\
& & Chamfer
& 0.035 & 0.020 & 0.018 & 0.040 & 0.099 & 0.037 & 0.086 & 0.048 \\
& & ATE
& 0.035 & 0.026 & 0.014 & 0.093 & 0.144 & 0.052 & 0.020 & 0.055 \\

\cmidrule(lr){2-11}

& \multirow{4}{*}{AnyCam~\cite{wimbauer2025anycam}}
& Accuracy
& 0.205 & 0.100 & 0.246 & 0.149 & 0.205 & 0.181 & 0.268 & 0.193 \\
& & Completeness
& 0.079 & 0.026 & 0.029 & 0.061 & 0.052 & 0.049 & 0.187 & 0.069 \\
& & Chamfer
& 0.142 & 0.063 & 0.138 & 0.105 & 0.129 & 0.115 & 0.228 & 0.131 \\
& & ATE
& 0.106 & 0.084 & 0.155 & 0.211 & 0.239 & 0.115 & 0.085 & 0.142 \\

\cmidrule(lr){2-11}

& \multirow{4}{*}{Ours (photo. only)}
& Accuracy
& 0.018 & 0.013 & 0.010 & 0.024 & 0.069 & 0.035 & 0.445 & 0.088 \\
& & Completeness
& 0.059 & 0.036 & 0.030 & 0.061 & 0.089 & 0.046 & 0.285 & 0.087 \\
& & Chamfer
& 0.038 & 0.025 & 0.022 & 0.043 & 0.079 & 0.041 & 0.365 & 0.088 \\
& & ATE
& 0.035 & 0.024 & 0.013 & 0.101 & 0.153 & 0.051 & 0.239 & 0.088 \\

\cmidrule(lr){2-11}

& \multirow{4}{*}{Ours (SIFT + NN)}
& Accuracy
& 0.017 & 0.015 & 0.014 & 0.025 & 0.071 & 0.033 & 0.023 & 0.028 \\
& & Completeness
& 0.063 & 0.042 & 0.028 & 0.067 & 0.094 & 0.045 & 0.129 & 0.067 \\
& & Chamfer
& 0.041 & 0.029 & 0.018 & 0.046 & 0.084 & 0.040 & 0.078 & 0.048 \\
& & ATE
& 0.036 & 0.024 & 0.013 & 0.094 & 0.146 & 0.051 & 0.023 & 0.055 \\

\cmidrule(lr){2-11}

& \multirow{4}{*}{Ours}
& Accuracy
& 0.018 & 0.013 & 0.010 & 0.022 & 0.068 & 0.034 & 0.026 & 0.027 \\
& & Completeness
& 0.060 & 0.034 & 0.028 & 0.061 & 0.095 & 0.049 & 0.128 & 0.065 \\
& & Chamfer
& 0.038 & 0.024 & 0.019 & 0.042 & 0.081 & 0.042 & 0.077 & 0.046 \\
& & ATE
& 0.035 & 0.024 & 0.013 & 0.096 & 0.146 & 0.051 & 0.023 & 0.055 \\

\bottomrule
\end{tabular}%
}
\end{table*}

\begin{figure*}[tbp]
    \centering
    \setlength{\tabcolsep}{1pt}
    \renewcommand{\arraystretch}{0.5} 

    \begin{tabular}{cccc}
        &
        \scriptsize \textbf{DroidCalib} &
        \scriptsize \textbf{photo. only} &
        \scriptsize \textbf{Ours} \\
        \noalign{\vspace{2pt}}
        
        \raisebox{-0.5\height}{\rotatebox{90}{\scriptsize \textbf{office}}} &
        \raisebox{-0.5\height}{\includegraphics[width=0.3\linewidth, trim={0cm 0cm 0cm 0cm}, clip]{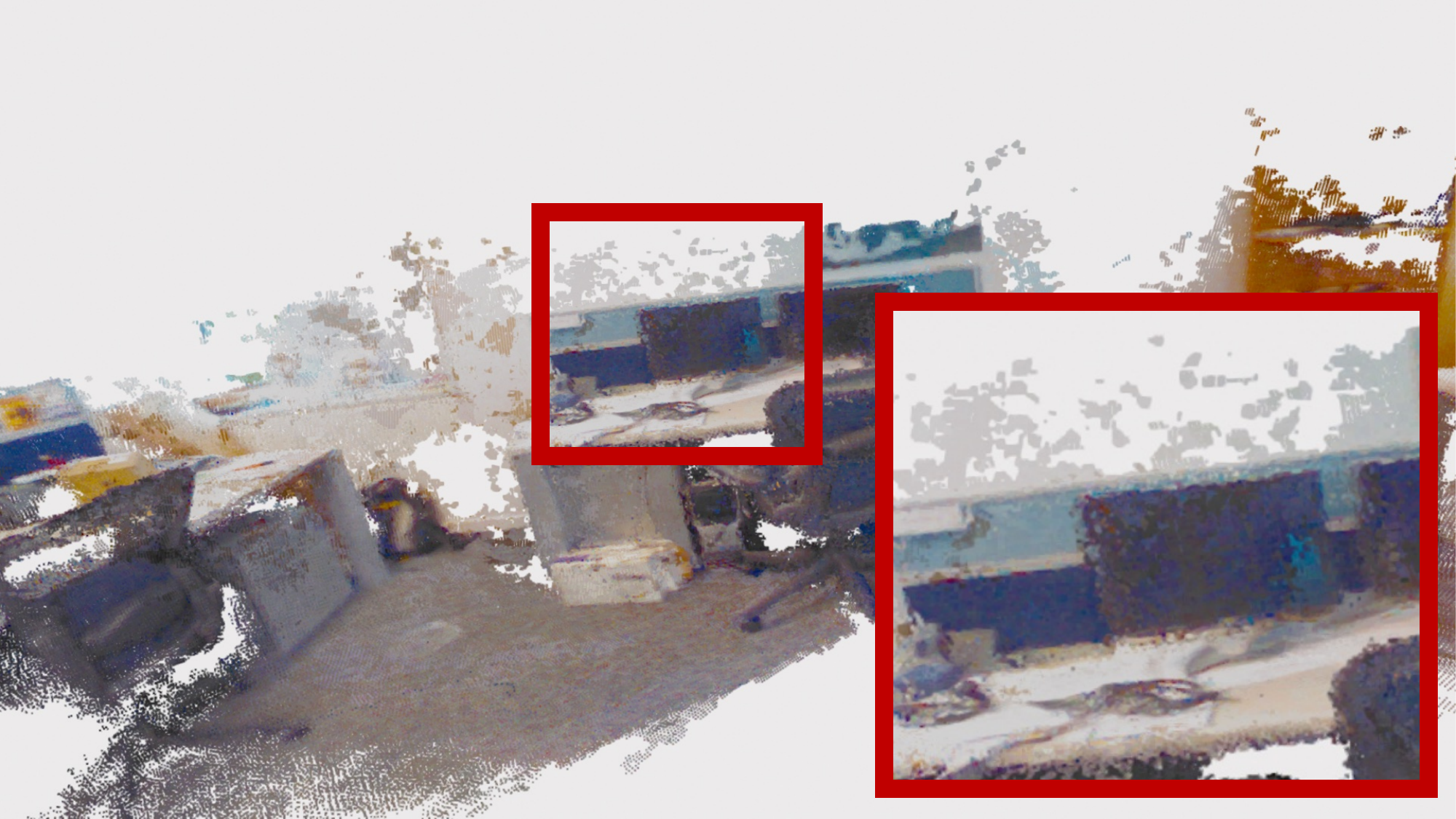}} &
        \raisebox{-0.5\height}{\includegraphics[width=0.3\linewidth, trim={0cm 0cm 0cm 0cm}, clip]{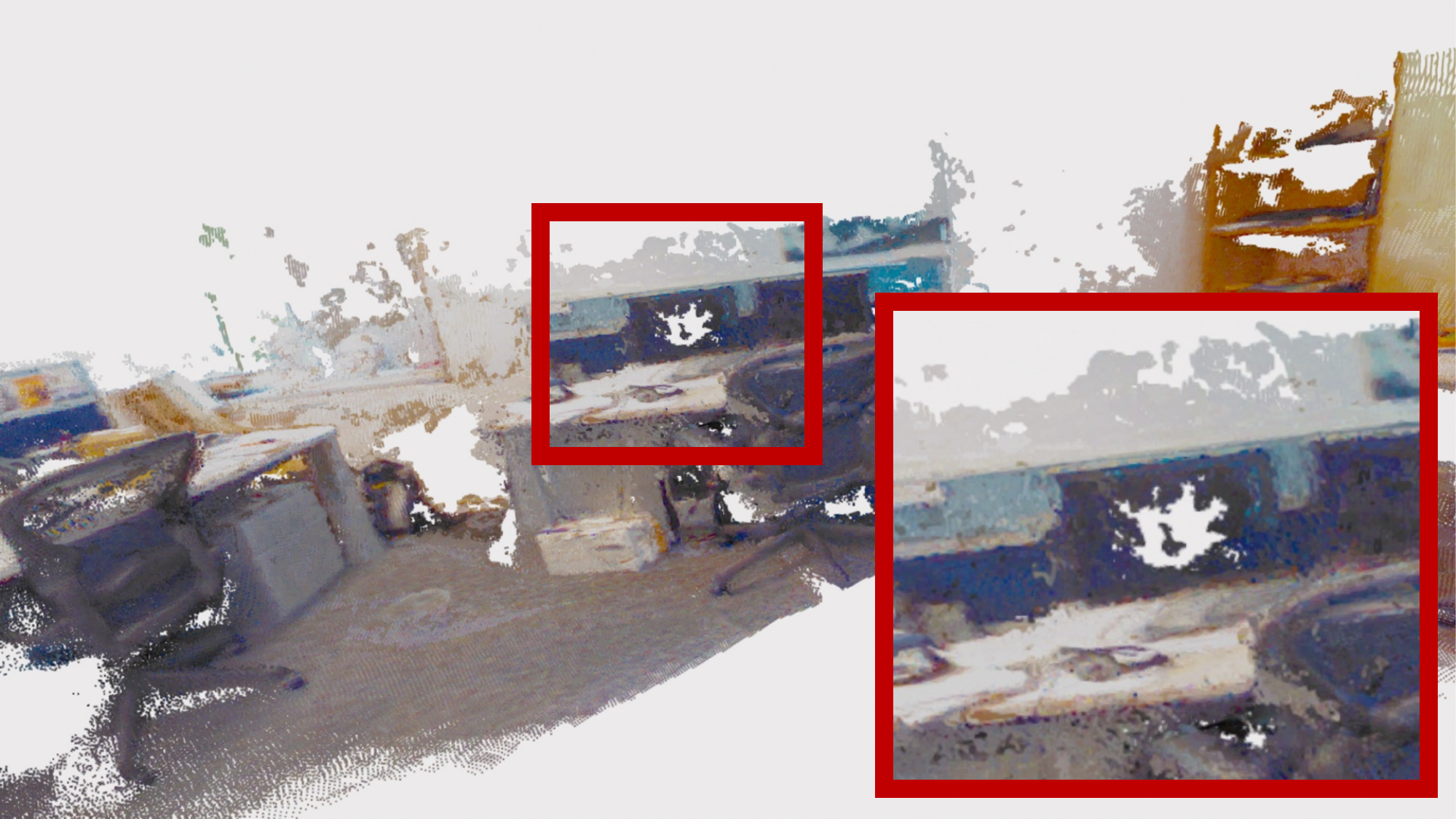}} &
        \raisebox{-0.5\height}{\includegraphics[width=0.3\linewidth, trim={0cm 0cm 0cm 0cm}, clip]{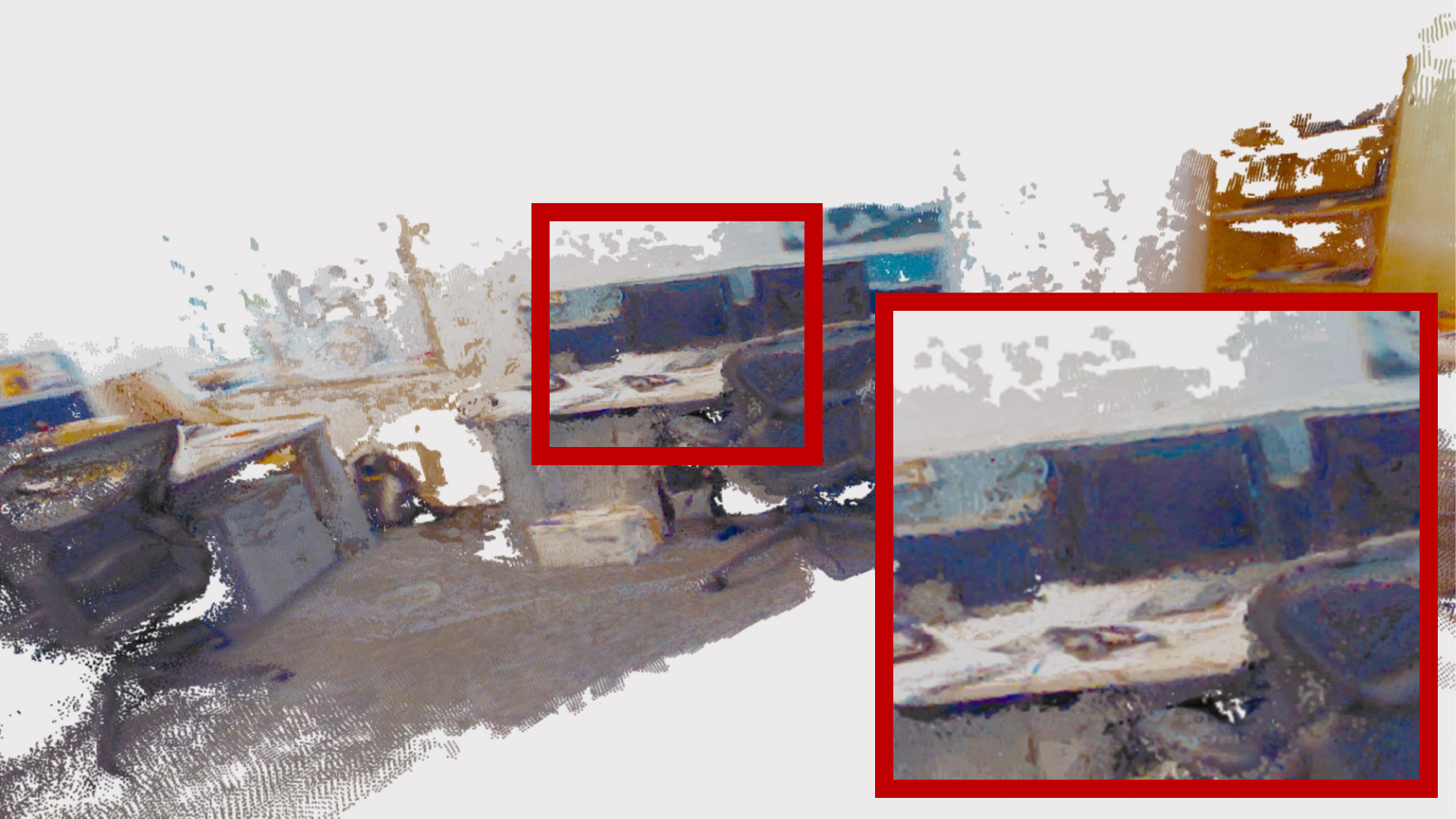}} \\
        \noalign{\vspace{2pt}}
        
        \raisebox{-0.5\height}{\rotatebox{90}{\scriptsize \textbf{fire}}} &
        \raisebox{-0.5\height}{\includegraphics[width=0.3\linewidth, trim={0cm 0cm 0cm 0cm}, clip]{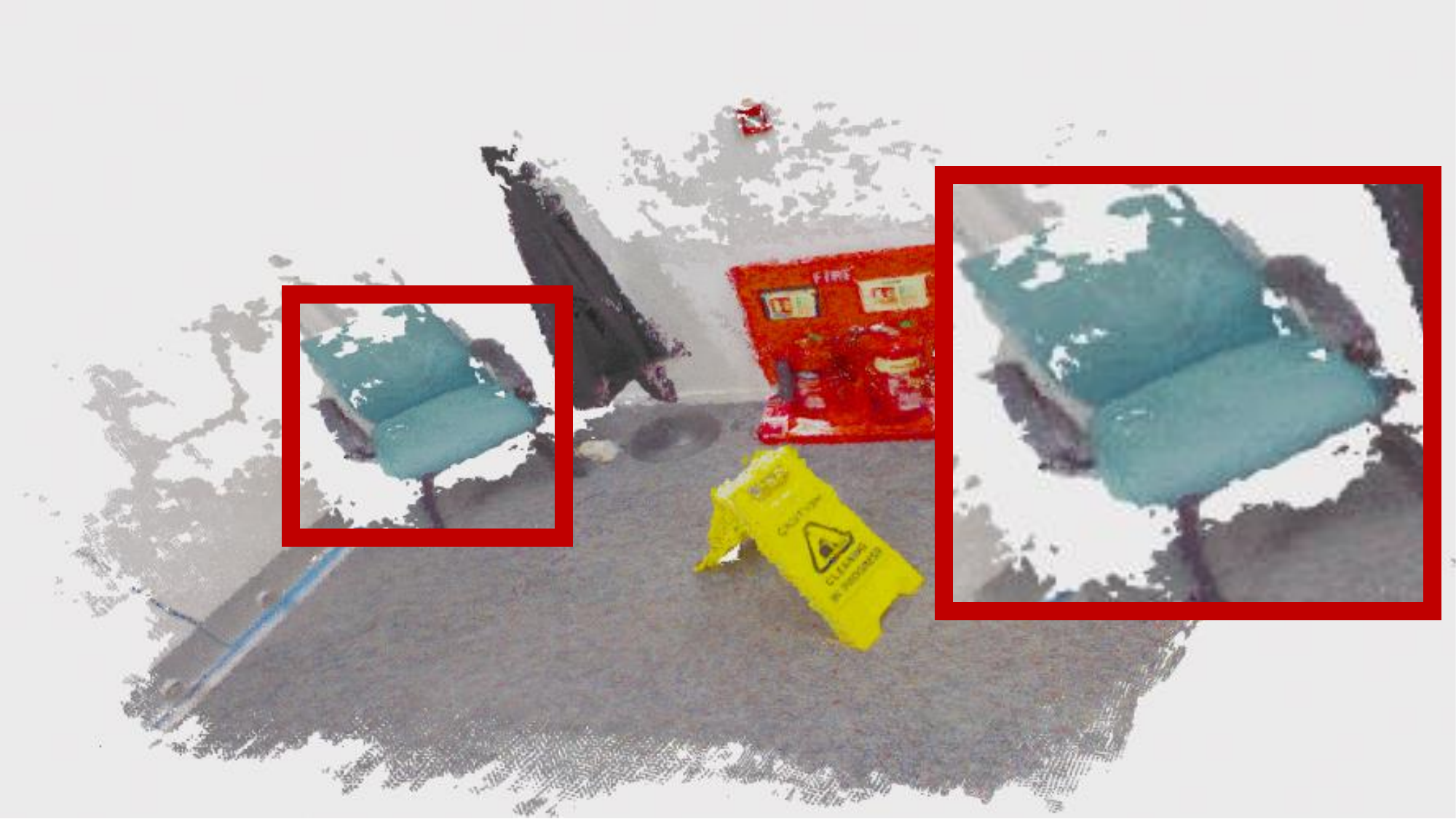}} &
        \raisebox{-0.5\height}{\includegraphics[width=0.3\linewidth, trim={0cm 0cm 0cm 0cm}, clip]{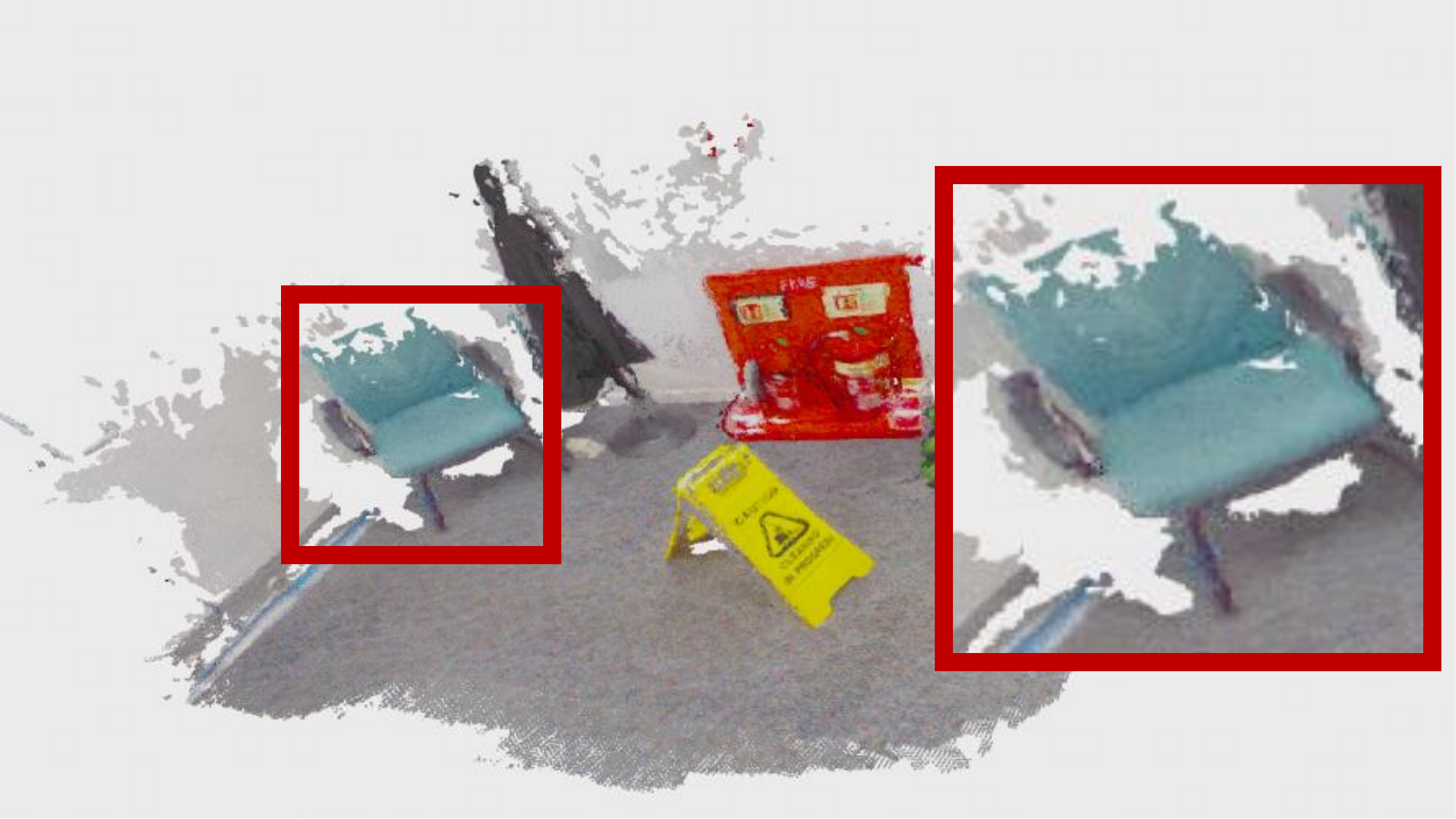}} &
        \raisebox{-0.5\height}{\includegraphics[width=0.3\linewidth, trim={0cm 0cm 0cm 0cm}, clip]{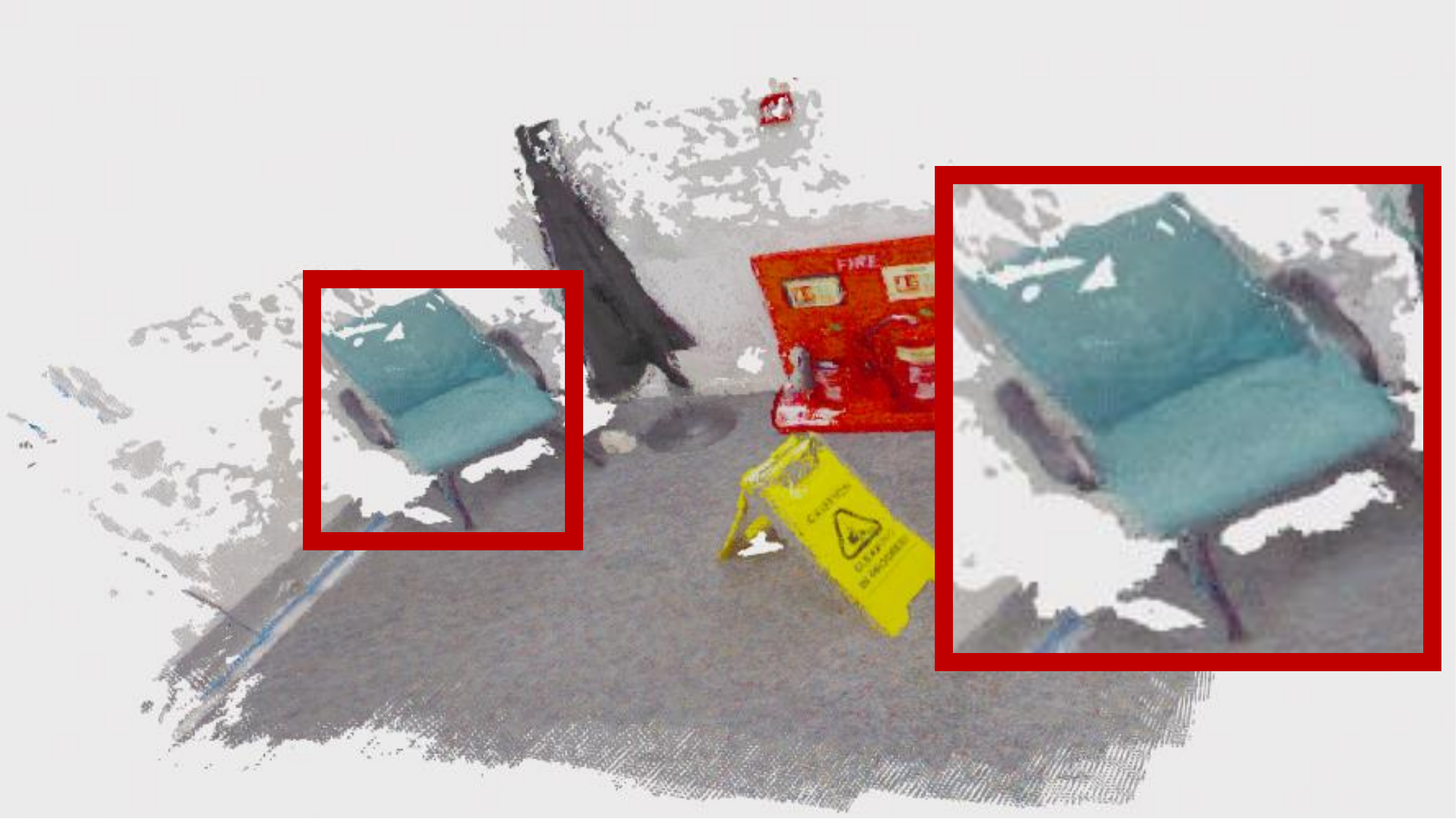}} \\
        \noalign{\vspace{2pt}}

        \raisebox{-0.5\height}{\rotatebox{90}{\scriptsize \textbf{MH000}}} &
        \raisebox{-0.5\height}{\includegraphics[width=0.3\linewidth, trim={0cm 0cm 0cm 0cm}, clip]{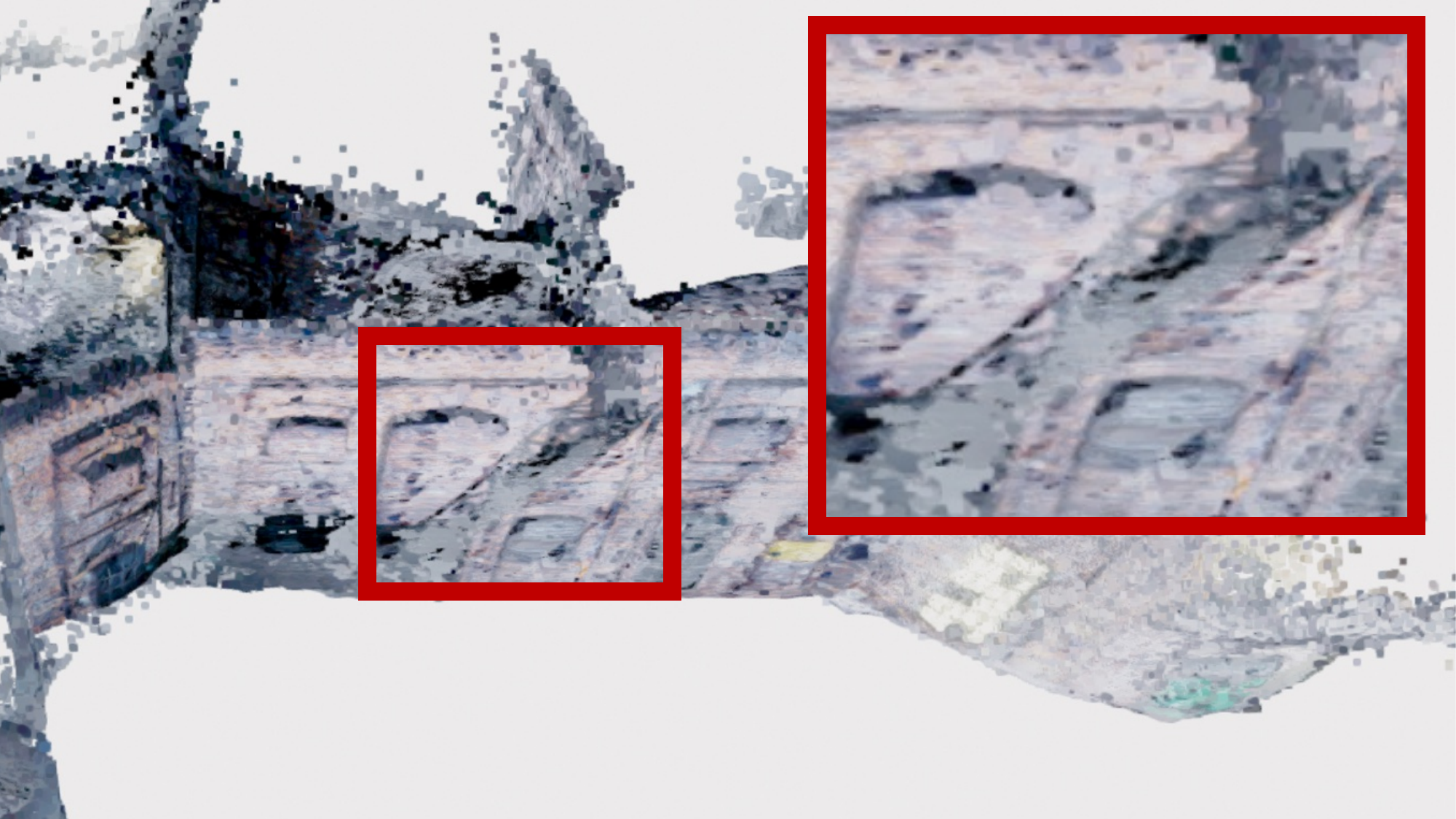}} &
        \raisebox{-0.5\height}{\includegraphics[width=0.3\linewidth, trim={0cm 0cm 0cm 0cm}, clip]{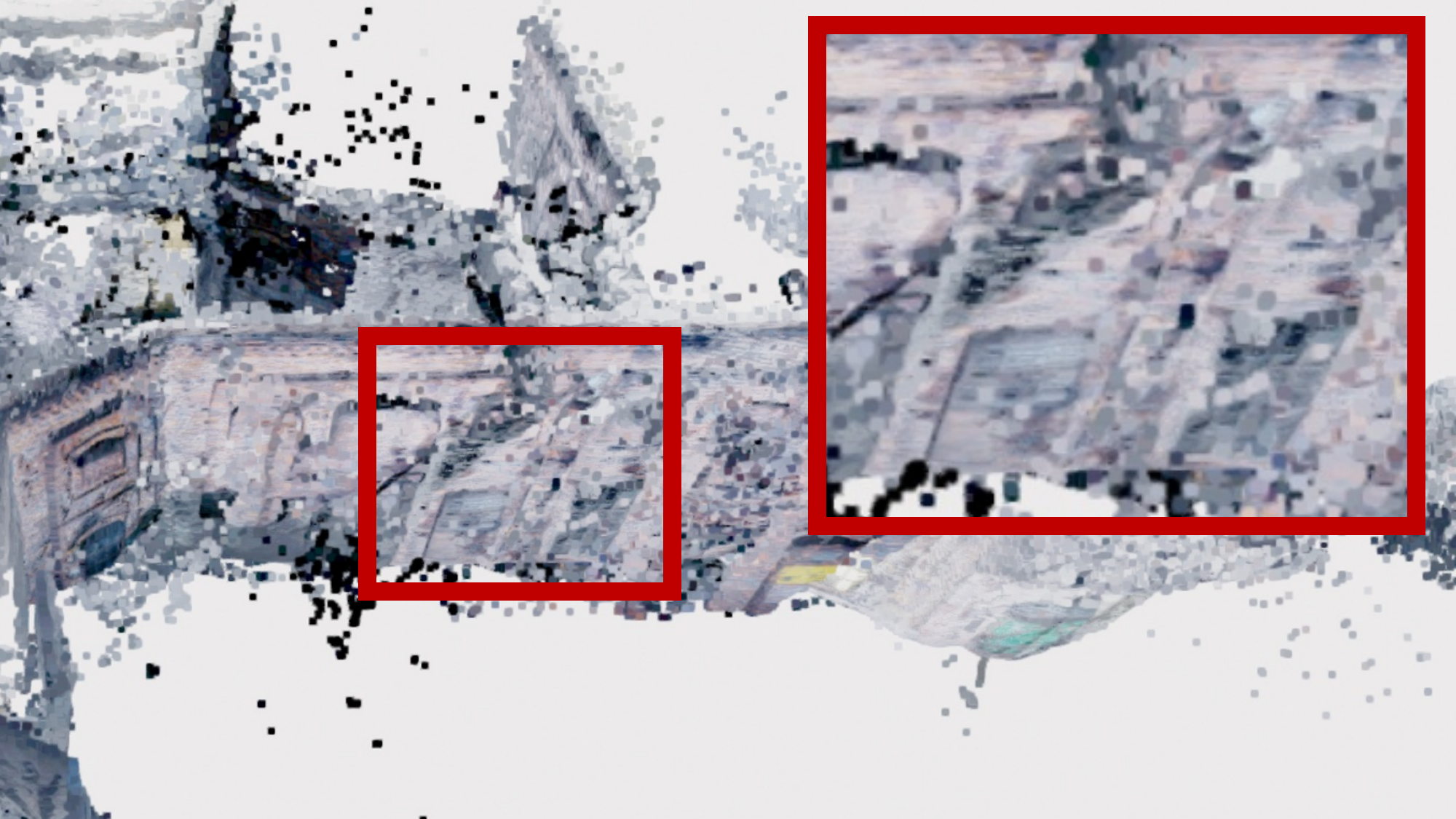}} &
        \raisebox{-0.5\height}{\includegraphics[width=0.3\linewidth, trim={0cm 0cm 0cm 0cm}, clip]{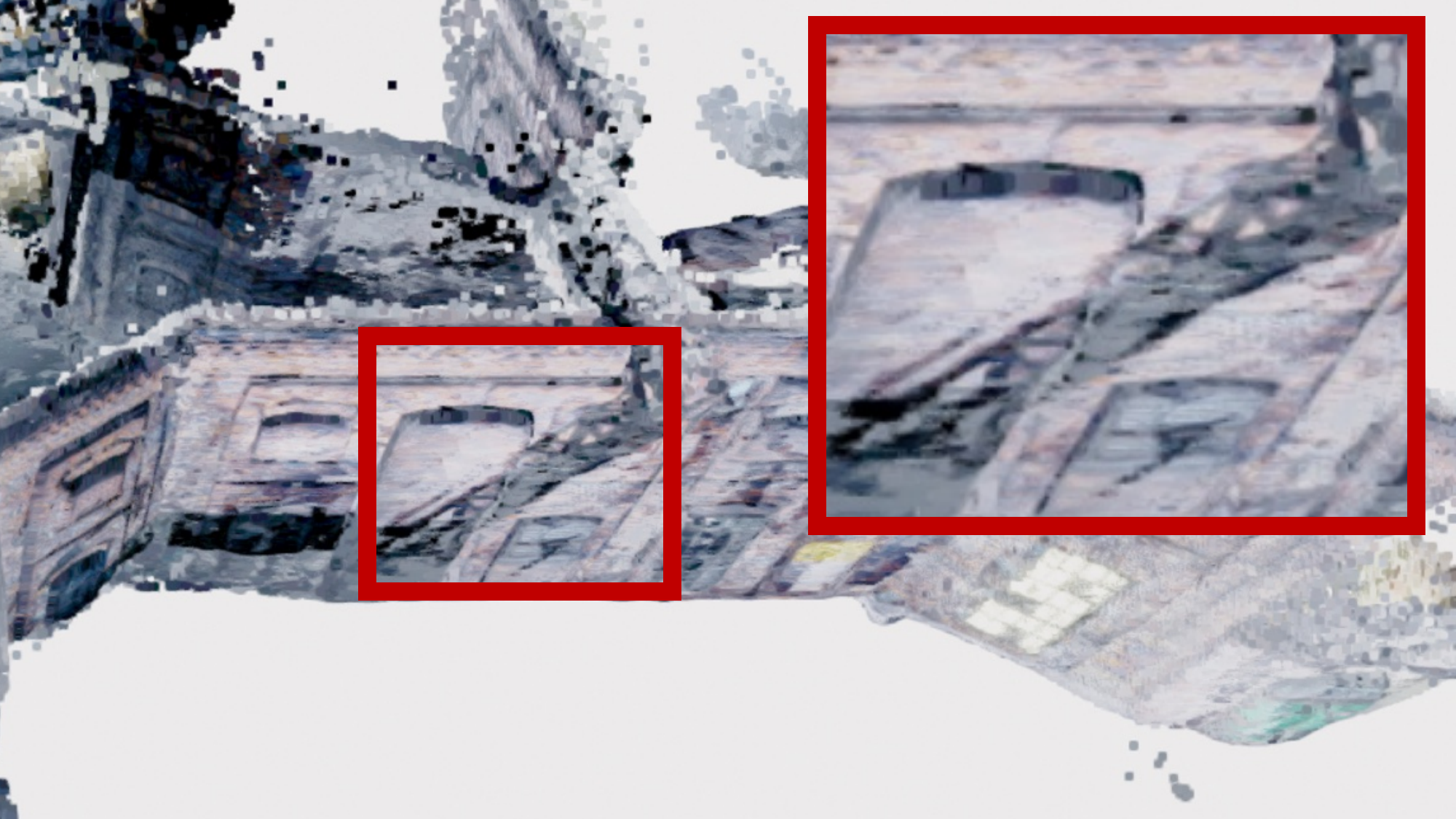}} \\
    \end{tabular}
    \caption{\textbf{Additional reconstruction results on 7-Scenes~\cite{shotton20137scenes} (\texttt{office}, \texttt{fire}), and TartanAir~\cite{wang2020tartanair} (\texttt{MH000}).} Red boxes highlight regions in which our photometric-only baseline produces substantial reconstruction errors. In contrast, \Ourname preserves geometric consistency.}
    \label{fig:recon_additional}
\end{figure*}

\boldparagraph{Additional qualitative 3D reconstruction results} 
In \Cref{fig:recon_additional}, we present additional 3D reconstruction results
on 7-Scenes~\cite{shotton20137scenes}, and
TartanAir~\cite{wang2020tartanair}. Consistent with \Cref{fig:recon}, the
photometric-only baseline exhibits minor artifacts in simple scenes
but shows larger holes in more complex scenes such as 7-Scenes \texttt{office}.
In contrast, \Ourname preserves smooth, geometrically consistent reconstructions
with quality comparable to DroidCalib.

\subsection{Trajectory estimation on TartanAir} 

\begin{figure*}[t]
    \centering
    \setlength{\tabcolsep}{1pt}
    \renewcommand{\arraystretch}{0.2}

    \newcommand{\plotheighttop}{2.4cm}
    \newcommand{\plotheightbottom}{2.65cm}

    \scriptsize
    \resizebox{0.85\linewidth}{!}{%
    \begin{tabular}{ccccc}
        & \large\textbf{\hspace{25pt}DroidCalib}
        & \large\textbf{AnyCam}
        & \large\textbf{Ours photo. only}
        & \large\textbf{\hspace{-25pt}Ours} \\

        \raisebox{-0.3cm}{\rotatebox{90}{\large\textbf{MH000}}}\hspace{5pt} &
        \includegraphics[valign=c, height=\plotheighttop, trim={0cm 1cm 1cm 0cm}, clip]{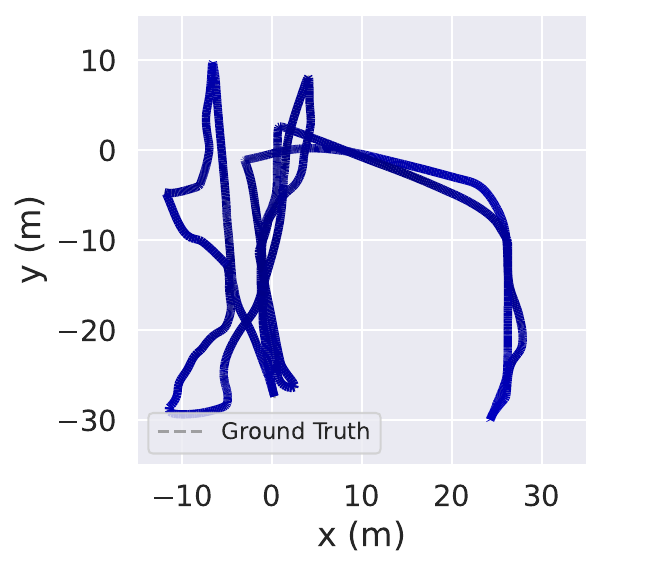} &
        \includegraphics[valign=c, height=\plotheighttop, trim={1.7cm 1cm 1cm 0cm}, clip]{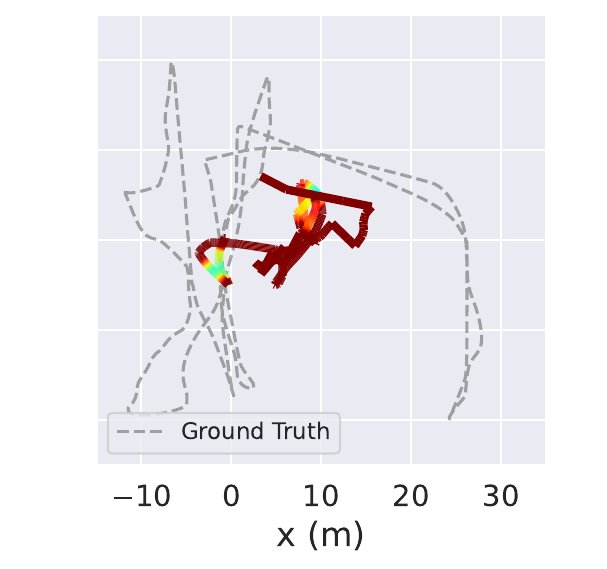} &
        \includegraphics[valign=c, height=\plotheighttop, trim={1.7cm 1cm 1cm 0cm}, clip]{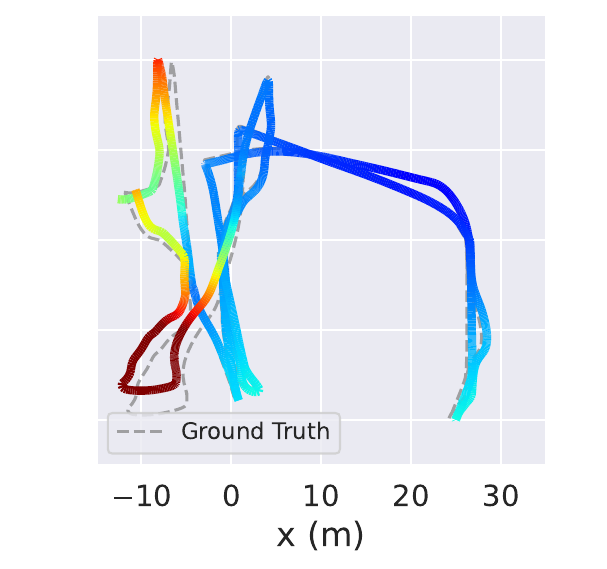} &
        \includegraphics[valign=c, height=\plotheighttop, trim={1.7cm 1cm -0.3cm 0cm}, clip]{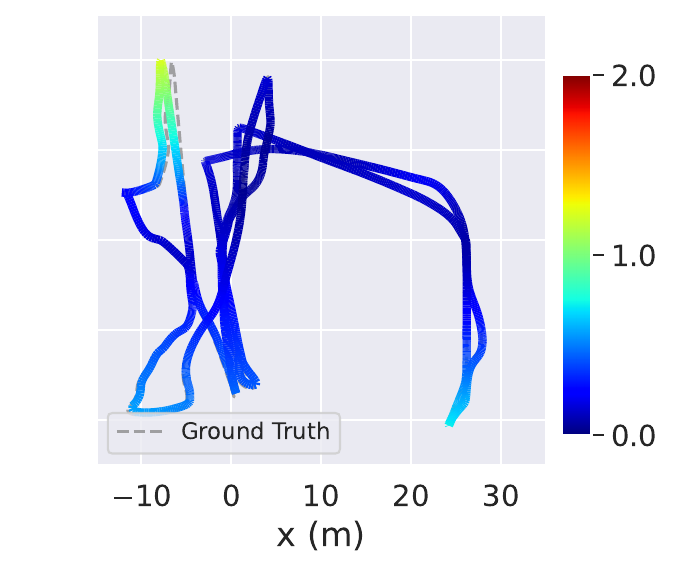} \\[-1pt]

        \raisebox{-0.3cm}{\rotatebox{90}{\large\textbf{MH006}}}\hspace{5pt} &
        \includegraphics[valign=c, height=\plotheighttop, trim={0cm 1cm 1cm 0cm}, clip]{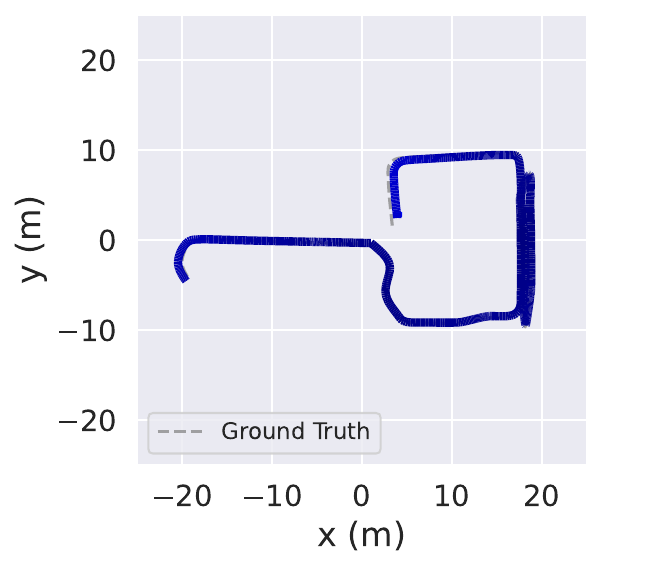} &
        \includegraphics[valign=c, height=\plotheighttop, trim={1.7cm 1cm 1cm 0cm}, clip]{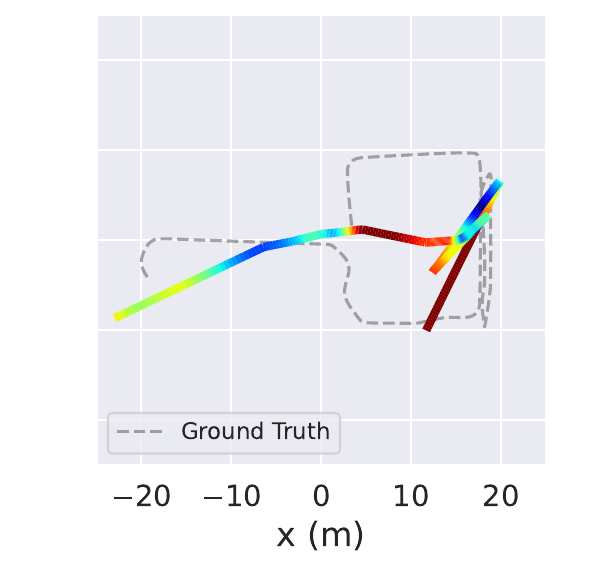} &
        \includegraphics[valign=c, height=\plotheighttop, trim={1.7cm 1cm 1cm 0cm}, clip]{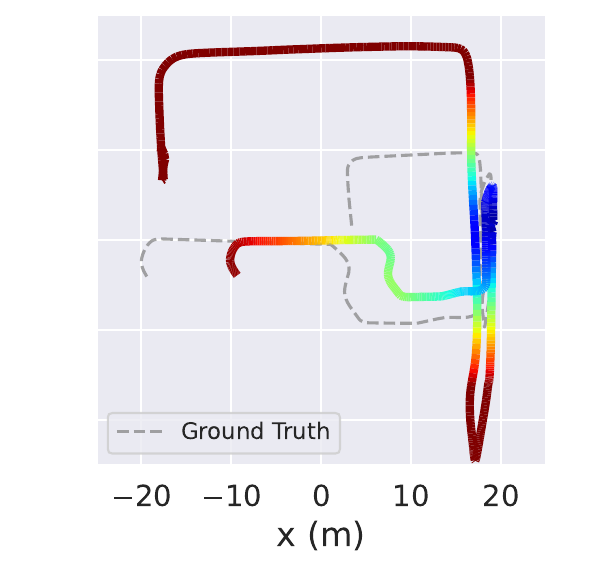} &
        \includegraphics[valign=c, height=\plotheighttop, trim={1.7cm 1cm 0cm 0cm}, clip]{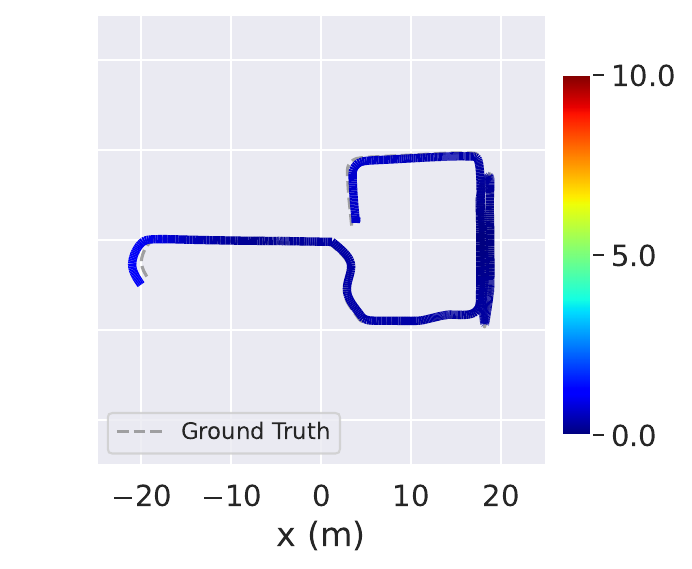} \\[-1pt]

        \raisebox{0.1cm}{\rotatebox{90}{\large\textbf{MH007}}}\hspace{5pt} &
        \includegraphics[valign=c, height=\plotheightbottom, trim={0cm 0cm 1cm 0cm}, clip]{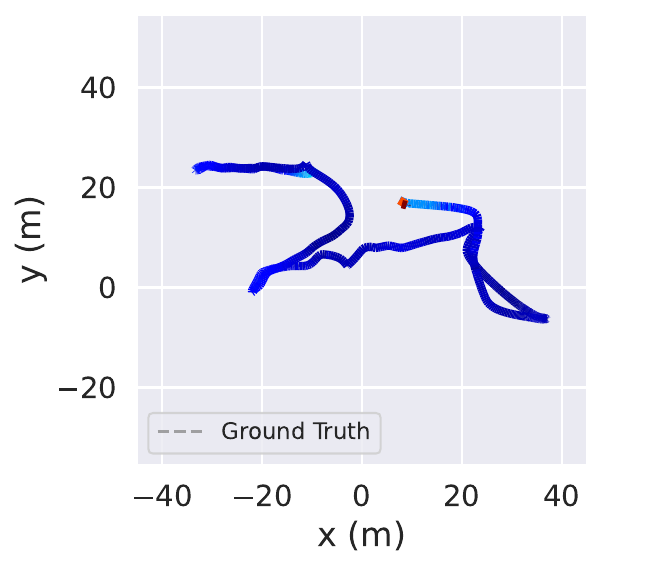} &
        \includegraphics[valign=c, height=\plotheightbottom, trim={1.7cm 0cm 1cm 0cm}, clip]{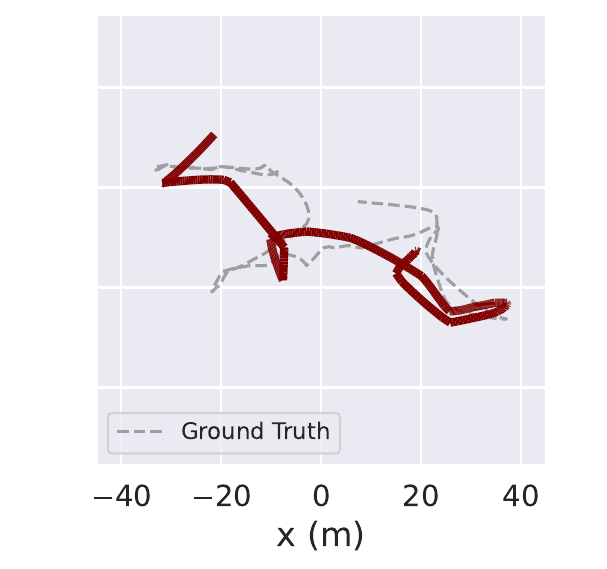} &
        \includegraphics[valign=c, height=\plotheightbottom, trim={1.7cm 0cm 1cm 0cm}, clip]{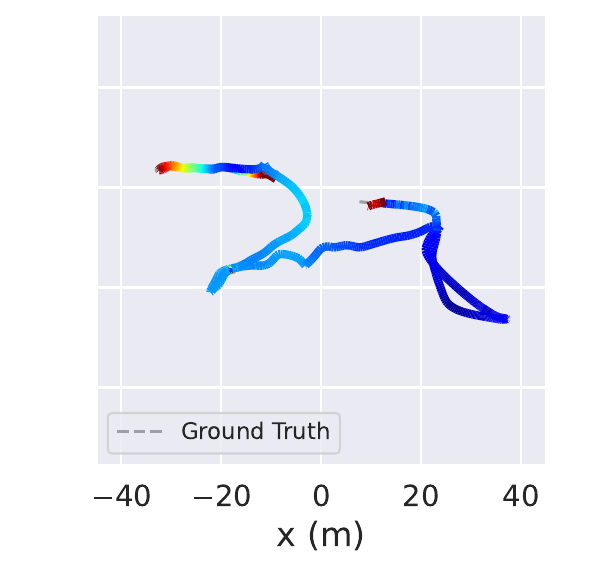} &
        \includegraphics[valign=c, height=\plotheightbottom, trim={1.7cm 0cm -0.3cm 0cm}, clip]{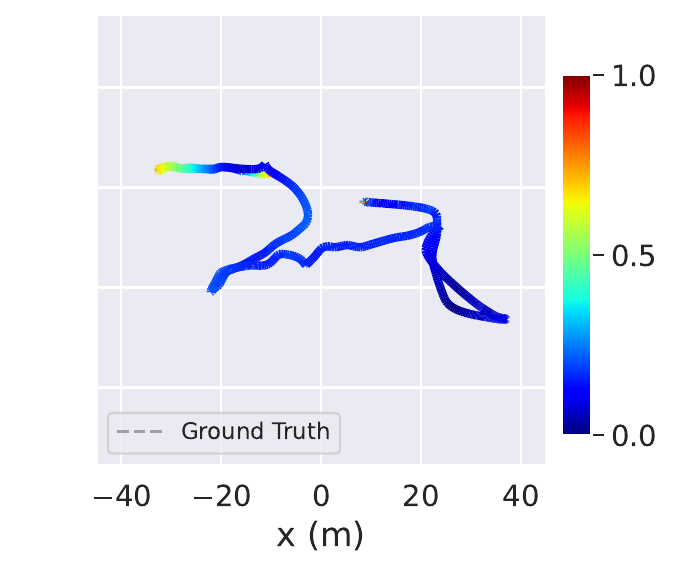} \\

    \end{tabular}%
    }

    \caption{\textbf{Trajectory estimation results on TartanAir~\cite{wang2020tartanair}.}
    Trajectory color encodes the ATE RMSE, transitioning from blue (low error) to red (high error). Our method demonstrates stability on challenging scenes while AnyCam and the photometric-only baseline fail.}
    \label{fig:additional_trajectory_results}
\end{figure*}

Complementing the quantitative ATE results presented in the main paper,
\Cref{fig:additional_trajectory_results} provides qualitative trajectory
estimates for three challenging, long-range sequences from the
TartanAir~\cite{wang2020tartanair} dataset: \texttt{MH000} (top), \texttt{MH006}
(middle), and \texttt{MH007} (bottom). We observe that
AnyCam~\cite{wimbauer2025anycam} fails to reliably track these complex
sequences, quickly losing the ground-truth path and accumulating massive errors.
Similarly, the photometric-only baseline proves insufficiently robust for
long-distance navigation; it suffers from severe drift in \texttt{MH000},
completely diverges from the rectangular loop path in \texttt{MH006}, and
accumulates noticeable positional error toward the end of \texttt{MH007}. In
contrast, \Ourname maintains highly robust tracking throughout all sequences. As
visually confirmed by the color-coded error magnitudes, our approach
consistently yields low-error estimates (dark blue) that tightly follow the
ground truth, corroborating that our method approaches the performance of the
densely supervised DroidCalib~\cite{hagemann2023droidcalib} on TartanAir in a
strictly zero-shot setting.

\subsection{Comparison to E-RayZer} \label{sec:supp__e-rayzer}

We compare \Ourname against E-RayZer~\cite{zhao2025erayzer}, a state-of-the-art
self-super\-vised novel view synthesis model. Unlike our direct warping
formulation, E-RayZer achieves photometric self-supervision through
differentiable 3D Gaussian Splatting. Because E-RayZer lacks a sequential SLAM
frontend, it cannot operate on the full sequence as other methods in
\Cref{tab:tum_results}. Therefore, we adopt a modified evaluation protocol on
the TUM-RGBD~\cite{sturm2012ate} benchmark to ensure a fair comparison between
E-RayZer and \Ourname. Specifically, we evaluate both methods on the first 200
frames of each scene: E-RayZer processes a sparse subset of 10 frames sampled
with a stride of 20, whereas our system processes the full continuous sequence
using standard keyframe selection. Furthermore, since E-RayZer predicts camera
intrinsics, we evaluate \Ourname in its self-calibrating mode. As shown in
\Cref{tab:e-rayzer}, our method demonstrates superior accuracy, reducing the
average ATE RMSE by $45\%$ compared to E-RayZer.

\begin{table*}[t]
\centering
\caption{\textbf{Quantitative comparison with E-RayZer~\cite{zhao2025erayzer}.} Evaluating ATE RMSE [m] on the first 200 frames of TUM-RGBD (Freiburg 1), \Ourname outperforms the modern Gaussian Splatting-based self-supervised model.}
\vspace{0.5em}
\label{tab:e-rayzer}

\renewcommand{\arraystretch}{0.9}

\resizebox{0.9\textwidth}{!}{%
\setlength{\aboverulesep}{1pt}
\setlength{\belowrulesep}{1pt}
\footnotesize
\setlength{\tabcolsep}{5pt}

\begin{tabular}{lcccccccccc}
\toprule

\textbf{Method}
& \texttt{360}
& \texttt{desk}
& \texttt{desk2}
& \texttt{floor}
& \texttt{plant}
& \texttt{room}
& \texttt{rpy}
& \texttt{teddy}
& \texttt{xyz}
& \textbf{Avg.} \\

\midrule

E-RayZer~\cite{zhao2025erayzer}
& \textbf{0.079}
& 0.148
& 0.206
& 0.053
& \textbf{0.030}
& 0.086
& \textbf{0.028}
& 0.042
& 0.012
& 0.076 \\

Ours
& 0.118
& \textbf{0.017}
& \textbf{0.036}
& \textbf{0.018}
& 0.092
& \textbf{0.034}
& 0.033
& \textbf{0.017}
& \textbf{0.011}
& \textbf{0.042} \\

\bottomrule
\end{tabular}%
}
\end{table*}

\subsection{Fine-tuning AnyCam on TartanAir} 

In \Cref{sec:experiments}, we compare \Ourname with the self-supervised visual
odometry framework AnyCam~\cite{wimbauer2025anycam}. Unlike \Ourname and
D$^3$FlowSLAM~\cite{yu2022d3flowslam}, AnyCam is trained on YouTube and casually
captured videos, resulting in a different training distribution. To reduce this
discrepancy, we fine-tune the released AnyCam model on the same
TartanAir~\cite{wang2020tartanair} training split used for Spade-SLAM for 40K
optimizer steps, using a reduced learning rate of $1 \times 10^{-5}$. As shown
in \Cref{tab:anycam_finetune}, fine-tuning substantially improves AnyCam, but a
large performance gap to Spade-SLAM remains.

\begin{table}[t]
\centering
\caption{\textbf{Fine-tuned AnyCam on TartanAir~\cite{wang2020tartanair} and 7-Scenes~\cite{shotton20137scenes}.} We report average ATE RMSE [m].}
\label{tab:anycam_finetune}
\begin{tabular}{lcc}
\toprule
\textbf{Method}
& TartanAir
& 7-Scenes \\
\midrule
AnyCam
& 7.94
& 0.207 \\
AnyCam (fine-tuned)
& 5.14
& 0.167 \\
\textbf{Ours}
& \textbf{0.50}
& \textbf{0.061} \\
\bottomrule
\end{tabular}
\vspace{-1em}
\end{table}

\section{Implementation details}
\label{sec:supp_implementation_details}

\subsection{Training dataset}

TartanAir is a large-scale, photorealistic synthetic benchmark designed
to challenge visual SLAM algorithms under extreme conditions. It comprises over
1000 trajectories across 30 diverse environments, including structured urban
centers, domestic indoor spaces, and unstructured natural landscapes. A key
feature of the dataset is the simulation of various drone flights, which provide
complex 6-DoF motion patterns ranging from smooth camera paths to aggressive
trajectories. The dataset includes significant environmental variations such as
alternating day and night cycles, rapidly changing illumination, and adverse
weather effects like rain, snow, and fog. These features make our SLAM system
robust to both perceptual and motion-based degradations.

\begin{figure*}[t]
    \centering
    \begin{subfigure}[t]{0.49\textwidth}
        \centering
        \includegraphics[width=\textwidth]{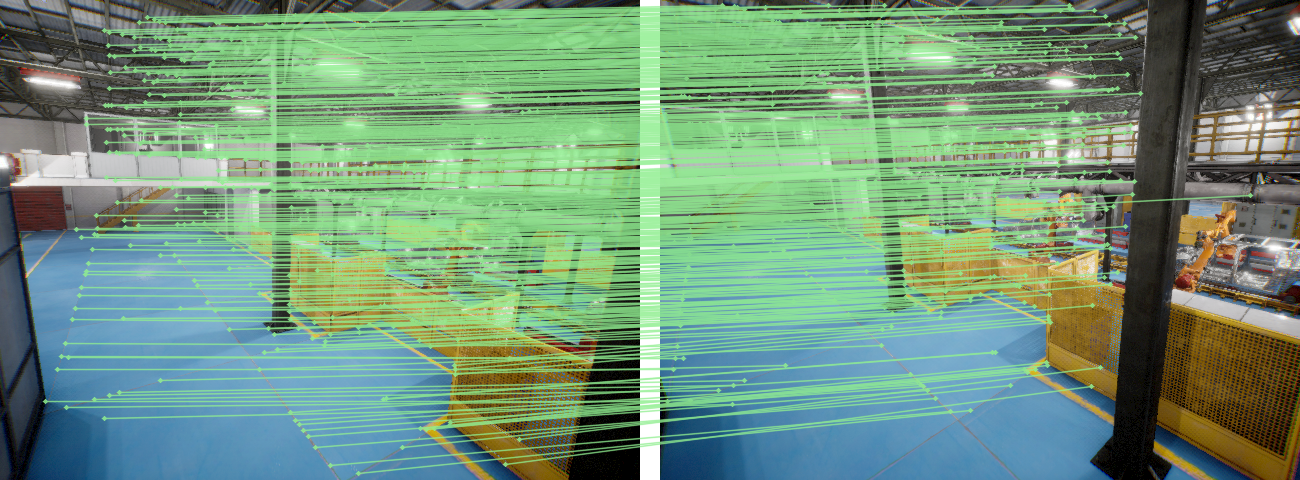}
        \label{fig:match_factory}
    \end{subfigure}
    \hfill
    \begin{subfigure}[t]{0.49\textwidth}
        \centering
        \includegraphics[width=\textwidth]{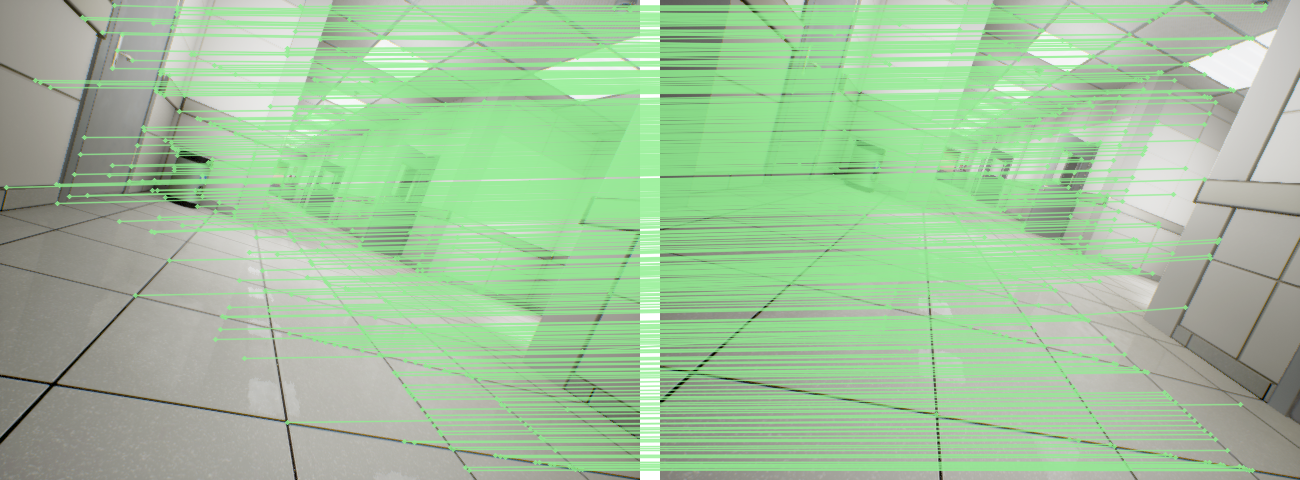}
        \label{fig:match_hospital}
    \end{subfigure}

    \vspace{-2mm} %

    \begin{subfigure}[t]{0.49\textwidth}
        \centering
        \includegraphics[width=\textwidth]{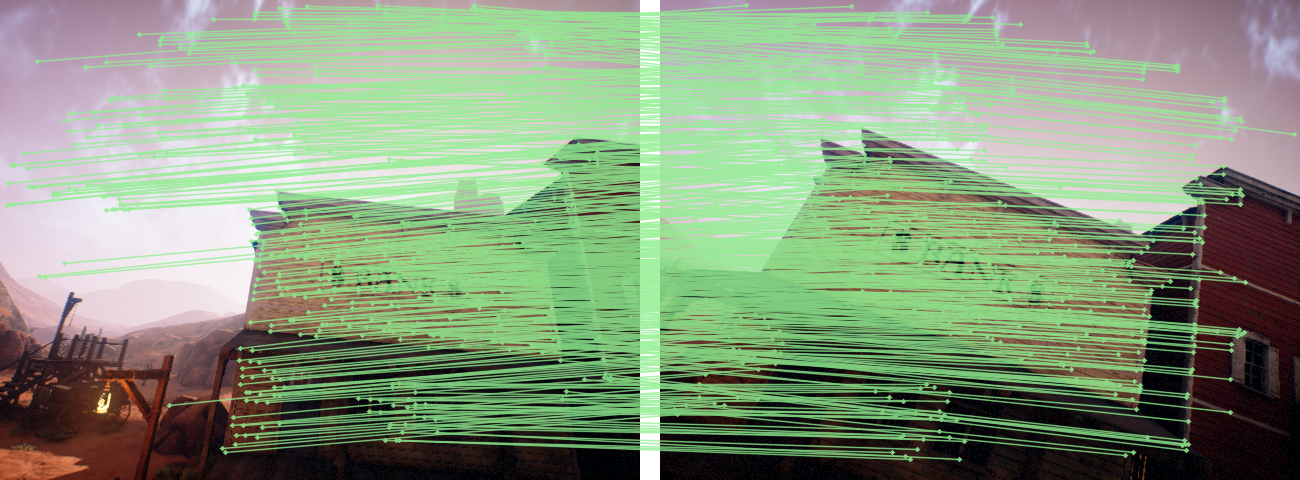}
        \label{fig:match_desert}
    \end{subfigure}
    \hfill
    \begin{subfigure}[t]{0.49\textwidth}
        \centering
        \includegraphics[width=\textwidth]{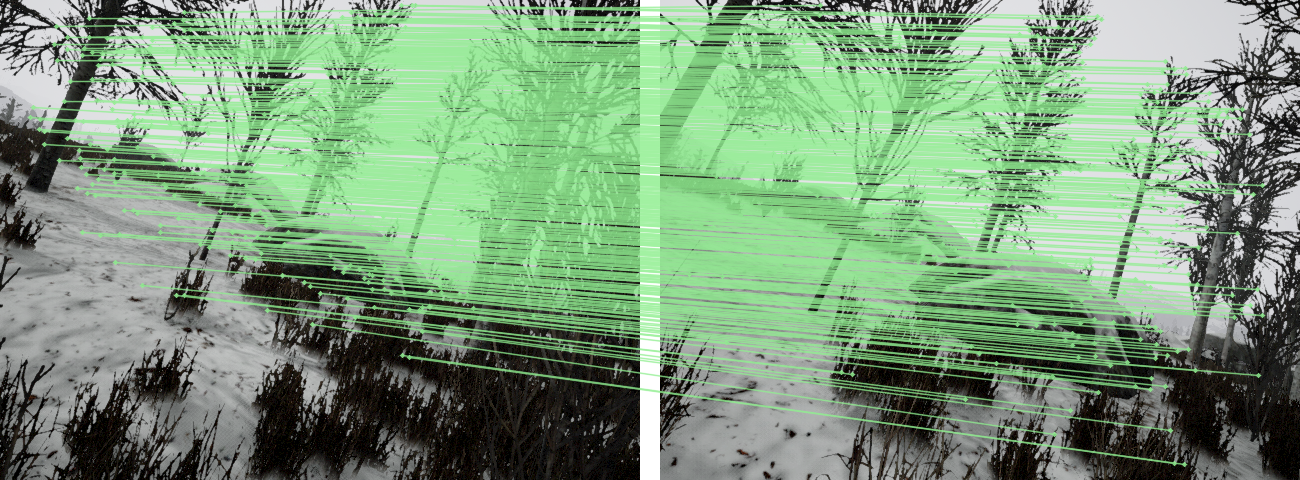} %
        \label{fig:match_extra}
    \end{subfigure}
    
    \caption{\textbf{Illustration of sparse correspondences used for training on TartanAir~\cite{wang2020tartanair}.}}
    \label{fig:sparse_correspondences_grid}
\end{figure*}

Figure \ref{fig:sparse_correspondences_grid} visualizes sparse correspondences obtained from LightGlue~\cite{lindenberger2023lightglue} and Superpoint~\cite{detone2018superpoint}
for our self-supervised training method across four distinct TartanAir
environments (from top-left to bottom-right: \texttt{carwelding},
\texttt{hospital}, \texttt{westerndesert}, and \texttt{seasonsforest\_winter}).
As shown, we obtain robust, well-distributed correspondences across diverse
conditions, demonstrating resilience to drastic lighting variations, indoor and
outdoor scene changes, and challenging textures.

\subsection{Adaptation and fine-tuning details}

For the domain adaptation and data scalability experiments in
\Cref{sec:adaptation}, we fine-tune our pretrained model using a reduced initial
learning rate of $1 \times 10^{-4}$ for 3K optimizer steps. Across all settings,
we use a batch size of 4 and process images at a resolution of $384 \times 512$
pixels, or at their native maximum resolution if lower. 

\subsection{SLAM system}

During inference, we deploy the trained network within the parallelized SLAM
architecture proposed in DROID-SLAM~\cite{teed2021DROID}. The system consists of
two asynchronous threads: a frontend that maintains a collection of recent
keyframes to estimate local poses and depth via the recurrent update operator,
and a backend that performs global bundle adjustment on the full frame graph to
rectify long-term drift.

To improve convergence stability in the self-supervised setting, an initial
warmup buffer of 30 keyframes is used. After tracking these frames, a factor
graph is initialized by connecting keyframes within a temporal radius of 3
(i.e., $|i-j| \le 3$), and bundle adjustment is run over all 30 keyframes to
ensure robust initialization. This warmup mechanism creates a densely
constrained graph prior to the first marginalization step, helping to compensate
for the higher uncertainty of self-supervised training signals. For a fair
comparison, all DROID-style systems evaluated in this paper, including
DROID-SLAM and DroidCalib~\cite{hagemann2023droidcalib}, use exactly the same
warmup and initialization protocol.

\boldparagraph{Inference speed}\label{sec:inference_speed}
Our method is real-time capable, maintaining high efficiency during operation.
Because we directly build upon the DROID-SLAM~\cite{teed2021DROID} backbone and
restrict our modifications entirely to the training pipeline, we achieve the
same inference speed as the original framework, typically ranging between 10 and
30 fps on 2 NVIDIA RTX 3090 GPUs depending on the input image resolution. At
inference time, our implementation leverages the optimized custom CUDA kernels
for DBA from prior works~\cite{teed2021DROID,hagemann2023droidcalib} to ensure
low-latency performance. For further details, we refer the reader to the
original DROID-SLAM paper.

\section{Limitations and future work}
\label{sec:supp_limit_future}

\subsection{Method limitations}
Our results highlight the complementary roles of sparse correspondence guidance
and dense photometric supervision: sparse matches enforce geometric consistency
and reduce drift, while photometric loss provides dense regularization across
the image domain. While we believe that Spade-SLAM presents an
innovative step forward in the area of self-supervised dense visual SLAM, there
are two limitations of our work.

\boldparagraph{Dependence on visual information}
Our method is fundamentally limited by the visual information available in the
input. Although sparse correspondences substantially improve training stability,
performance can degrade on long sequences dominated by textureless regions,
such as blank walls, where few reliable keypoints can be extracted. Performance
is also weaker in outdoor scenes containing large regions with few stable
correspondences, such as the sky or moving foliage, where establishing robust
geometric constraints remains challenging.

\boldparagraph{Dependence on matcher performance}
Our training framework relies on sparse correspondences as geometric
supervision, and its performance is therefore bounded by the accuracy and
spatial coverage of the underlying matcher. As shown in the ablation in
\Cref{tab:ablation}, training with sparse ground-truth correspondences
consistently outperforms training with correspondences obtained from a real
matcher. Consequently, improvements in sparse matching quality would
directly benefit the final tracking and reconstruction accuracy of our method.

\subsection{Research directions}

We identify several promising directions for extending the scope and robustness
of Spade-SLAM.

\boldparagraph{Feed-forward models}
A promising direction for future work is to extend our hybrid supervision to
feed-forward 3D reconstruction models, such as VGGT~\cite{wang2025vggt}, within
a self-supervised framework. In our current
DROID-based~\cite{teed2021DROID} backbone, the DBA layer provides implicit
geometric regularization by enforcing reprojection consistency during training.
In contrast, feed-forward models typically lack this iterative optimization
component. Investigating whether our supervision strategy can maintain its
efficacy in the absence of a DBA layer would be a critical test of its
robustness.

\boldparagraph{Dynamic objects}
Similar to most SLAM systems like our baseline DROID-SLAM~\cite{teed2021DROID},
our method is designed for static scenes. While small dynamic objects are
automatically filtered out through the confidence mechanism and
MAGSAC~\cite{barath2020magsac}, larger dynamic objects will likely lead to a
failure in camera tracking. In the future, it would be interesting to extend our
approach to enable training and inference on videos with more dynamic content.

\boldparagraph{Additional sensing modalities}
Spade-SLAM is designed as a visual-only SLAM system and achieves promising
results. To improve its performance even further, e.g. for challenging
environments such as VR headsets or drones, it would be interesting to
incorporate additional sensors such as an IMU.

\end{document}